\documentclass[10pt,twocolumn,letterpaper]{article}

\usepackage{cvpr}              % To produce the CAMERA-READY version
\usepackage[dvipsnames]{xcolor}
\renewcommand{\paragraph}[1]{\noindent\textbf{#1.}}

\def\method{\textsc{SpatialGen}\xspace}
\def\scene_number{12,328\xspace}
\def\room_number{57,431\xspace}
\def\image_number{4.7M\xspace}
\def\train_scene_number{57,381\xspace}
\def\test_scene_number{50\xspace}

\usepackage[ruled]{algorithm2e}
\usepackage{multirow}
\usepackage{soul}
\usepackage{stfloats}

\newcommand{\C}{\mathbf{C}}
\newcommand{\I}{\mathbf{I}}
\renewcommand{\S}{\mathbf{S}}
\renewcommand{\P}{\mathbf{P}}

\definecolor{cvprblue}{rgb}{0.21,0.49,0.74}
\usepackage[pagebackref,breaklinks,colorlinks,allcolors=cvprblue]{hyperref}
\usepackage[normalem]{ulem}
\usepackage[symbol]{footmisc}

\title{\method: Layout-guided 3D Indoor Scene Generation  }

\author{
Chuan Fang\textsuperscript{\rm 1}$^*$,
Heng Li\textsuperscript{\rm 1},
Yixun Liang\textsuperscript{\rm 1},
Jia Zheng\textsuperscript{\rm 2}, \\
Yongsen Mao\textsuperscript{\rm 2},
Yuan Liu\textsuperscript{\rm 1},
Rui Tang\textsuperscript{\rm 2},
Zihan Zhou\textsuperscript{\rm 2},
Ping Tan\textsuperscript{\rm 1} \\
\textsuperscript{\rm 1}Hong Kong University of Science and Technology,
\textsuperscript{\rm 2}Manycore Tech Inc.\\
{\tt\small \url{https://manycore-research.github.io/SpatialGen}}
}

\begin{document}
% \maketitlesupplementary

\twocolumn[{
\renewcommand\twocolumn[1][]{#1}%
\maketitle
\centering
\vspace{-5mm}
\includegraphics[width=\linewidth]{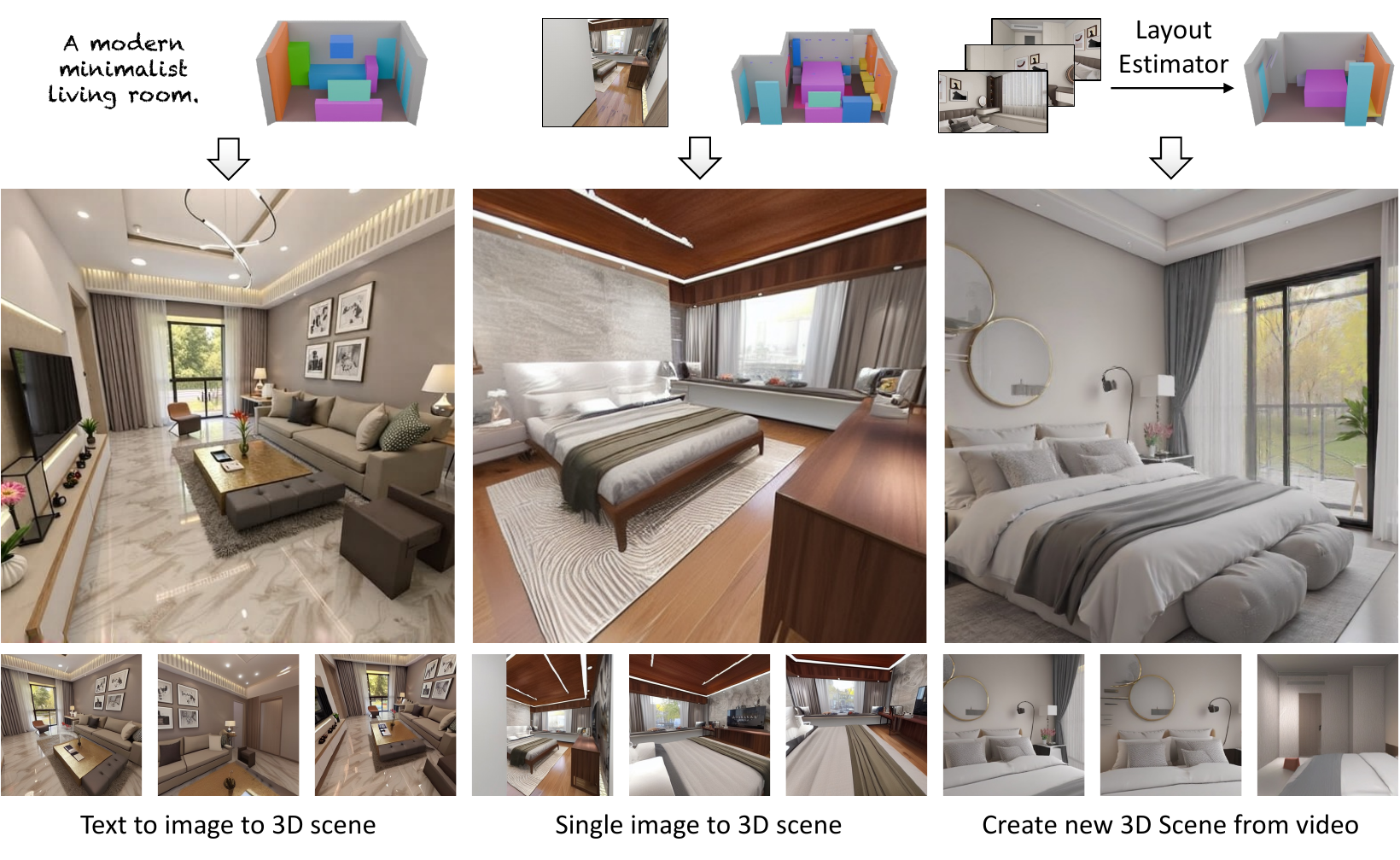}
\captionof{figure}{Given a 3D semantic layout, \method can generate a 3D indoor scene conditioned on either a textual description (left) or a reference image (middle). Furthermore, it can transform a real-world scene, where its 3D layout is estimated from a video by a layout estimator~\cite{SpatialLM}, into some brand new scenes.}
\vspace{2em}
\label{fig:teaser}
}]

\footnotetext[1]{Work done during an internship at Manycore Tech Inc.}
% \footnotetext[2]{Corresponding author.}

\begin{abstract}
Creating high-fidelity 3D models of indoor environments is essential for applications in design, virtual reality, and robotics. However, manual 3D modeling remains time-consuming and labor-intensive. While recent advances in generative AI have enabled automated scene synthesis, existing methods often face challenges in balancing visual quality, diversity, semantic consistency, and user control. A major bottleneck is the lack of a large-scale, high-quality dataset tailored to this task. To address this gap, we introduce a comprehensive synthetic dataset, featuring \scene_number structured annotated scenes with \room_number rooms, and \image_number photorealistic 2D renderings. Leveraging this dataset, we present SpatialGen, a novel multi-view multi-modal diffusion model that generates realistic and semantically consistent 3D indoor scenes. Given a 3D layout and a reference image (derived from a text prompt), our model synthesizes appearance (color image), geometry (scene coordinate map), and semantic (semantic segmentation map) from arbitrary viewpoints, while preserving spatial consistency across modalities. SpatialGen consistently generates superior results to previous methods in our experiments. We are open-sourcing our data and models to empower the community and advance the field of indoor scene understanding and generation.
\end{abstract}

\section{Introduction}
\label{sec:intro}

Indoor scene generation aims to produce spatially coherent and photorealistic 3D indoor environments. As a fundamental challenge in computer vision, this task underpins diverse applications, including immersive films and games, interior design, and augmented/virtual reality (AR/VR). Moreover, it also provides diverse and physically realistic environments in robotic simulation for training and evaluating robot navigation and interaction capabilities.

A major consideration in developing 3D scene generation methods is the trade-off between \emph{realism} and \emph{scene diversity}. Procedural modeling methods~\cite{InfinigenIndoors, MakeItHome, LayoutGPT} leverage hand-crafted heuristic rules and geometric constraints in the graphics engines, which produce highly realistic and physically plausible indoor environments. However, these scenes lack diversity. Recent 3D generative methods automatically generate scene layouts~\cite{ATISS, DiffuScene} or other 3D representations like NeRFs~\cite{GAUDI} and 3D Gaussians~\cite{Director3D}. But these methods exhibit limited layout and appearance realism, primarily due to the scarcity of annotated 3D data. In comparison, image-based methods utilize diffusion models to generate panoramas~\cite{MVDiffusion, ControlRoom3D} or multi-view images~\cite{CAT3D, Bolt3D} followed by 3D reconstruction. By leveraging powerful 2D priors, these methods show promise in striking a better balance between realism and scene diversity. Image-based methods, however, face additional challenges in multi-view \emph{semantic consistency}. While recent video generation methods~\cite{ViewCrafter, See3D, Gen3C} have improved temporal coherence, synthesizing semantically consistent content when exploring beyond input views remain highly challenging.

To this end, \textbf{3D semantic layout} prior (\cref{fig:teaser}) has been employed in the literature to guide the generation process. But due to the lack of a large-scale dataset with paired 3D layout and images (or videos), existing layout-conditioned methods resort to one of the following two strategies: \emph{score distillation}~\cite{SetTheScene, SceneCraft, Layout2Scene, GALA3D} and \emph{panorama-as-proxy}~\cite{ControlRoom3D, Ctrl-Room}. The former directly distills powerful 2D pre-trained models for 3D content creation, avoiding the need for large-scale training data. But due to the inherent limitation of the SDS method~\cite{DreamFusion}, results produced by these methods suffer from severe visual artifacts (\eg, over-saturation, lack of details). In contrast, the latter makes use of a special type of data, namely panoramas, for which large datasets with diverse scenes and annotations are available (\eg, Structured3D dataset~\cite{Structured3D}). However, since panorama images are captured at fixed camera locations, models trained on such data have limited ability to extrapolate to novel viewpoints, restricting their application in real-world tasks.

To overcome these limitations, we collect a new indoor scene dataset on a much larger scale. Our dataset features \image_number panoramic images with precise 2D and 3D layout annotations, spanning \room_number rooms and \scene_number scenes. With this dataset, we take a new approach to 3D scene synthesis by building a scalable multi-view diffusion (MVD) model conditioned on 3D layout priors, which achieves high semantic consistency while maintaining the realism and scene diversity in the results.

We introduce \method, a novel framework for high-fidelity 3D indoor scene generation from a 3D room layout. \emph{First}, we convert the 3D semantic layout into view-specific representations comprising coarse semantic maps and scene coordinate maps~\cite{SceneCoordRegression}. \emph{Second}, we design a layout-guided attention mechanism that alternatively operates through: (i) cross-view attention for consistent information propagation across different viewpoints; (ii) cross-modal attention for fine-grained feature alignment between appearance, semantic, and geometric representations. This mechanism enables the joint synthesis of photorealistic RGB images, precise object semantic maps, and accurate scene coordinates for both input and novel viewpoints. \emph{Finally}, we employ an iterative multi-view generation strategy to ensure complete scene coverage, followed by 3D Gaussian splatting optimization that reconstructs an explicit radiance field to enable free-viewpoint rendering.

\begin{table*}[t]
    \small
    \centering
    \renewcommand{\tabcolsep}{4pt}
    \caption{Statistics of the datasets for indoor scene generation. $^\dagger$: object annotations are provided by Ctrl-Room~\cite{Ctrl-Room}.}
    \vspace{-3mm}
    \begin{tabular}{l|ccccc|cc}
        \toprule
        \multirow{2}{*}{Dataset (year)} & \multirow{2}{*}{source} & \multirow{2}{*}{\#scenes} & \multirow{2}{*}{\#images} & \multirow{2}{*}{\#objects} & \multirow{2}{*}{image type} & \multicolumn{2}{c}{annotations} \tabularnewline
        & & & & & & layouts & objects \tabularnewline
        \midrule
        SUN R-GBD (2015) & real & - & 10.3K & 59K & perspective image & $\bullet$ & $\bullet$ \tabularnewline
        ScanNet (2017) & real & 1,513 & 2.5M & 36K & regular video & & $\bullet$  \tabularnewline
        Matterport3D (2017) & real & 90 & 10.8K & 41K & sparse panoramas & & $\bullet$ \tabularnewline
        ScanNet++ v2 (2024) & real & 1,006 & 11.1M & 111K & regular video & & $\bullet$ \tabularnewline
        \midrule
        Structured3D (2020) & syn. & 3,500 & 196.5K & 150K$^\dagger$ & panorama image & $\bullet$ & $\bullet^\dagger$ \tabularnewline
        Hypersim (2021) & syn. & 461 & 77.4K & 58K & regular video & & $\bullet$ \tabularnewline
        \method dataset (ours) & syn. & \textbf{\scene_number} & \textbf{\image_number} & \textbf{1M} & panoramic video & $\bullet$ & $\bullet$ \tabularnewline
        \bottomrule
    \end{tabular}
    \label{tab:dataset}
\end{table*}

Our main contributions are summarized as follows:
\begin{itemize}
    \item We introduce a new large-scale dataset featuring over \image_number panoramic images of \room_number rooms and precise 2D and 3D layout annotations. This dataset fills a critical gap in 3D scene modeling by providing comprehensive multi-view data with structural annotations.
    
    \item We present \method, a new framework for layout-guided indoor scene generation. At the core of this framework is a novel multi-view multi-modal image diffusion method conditioned on a given layout prior, which generates semantically and geometrically consistent images from arbitrary viewpoints.
    
    \item Extensive evaluations conducted on text or image to 3D scene generations demonstrate that our method generates substantially more realistic and plausible 3D scenes.
\end{itemize}

\section{Related Work}
\label{sec:related}

\paragraph{Procedural \& 3D-based Scene Generation} Procedural generation (PCG) methods~\cite{Infinigen, InfinigenIndoors} create 3D scenes with hand-crafted rules or constraints. Recent approaches integrate large language models (LLMs), either to generate scene layouts for subsequent object retrieval or shape synthesis~\cite{LayoutGPT, Holodeck, LayoutVLM}, or to act as agents that produce Python scripts controlling procedural frameworks~\cite{3D-GPT, SceneCraft2}.

3D-based methods generate 3D scene representations using generative models trained on datasets with 3D annotations. ATISS~\cite{ATISS} and DiffuScene~\cite{DiffuScene} predict compact layout parameters for scene objects. DiffInDScene~\cite{DiffInDScene}, PDD~\cite{PyramidDiffusion}, and SceneFactor~\cite{SceneFactor} introduce a semantic layout as an intermediate guide to generate the indoor scene with an explicit geometric representation. But the lack of annotated 3D scene datasets results in subpar performance and limited generalization of such methods.

%-------------------------------------------------------------------------
\paragraph{Image-based Scene Generation} In contrast, image-based methods exploit strong 2D priors in pretrained diffusion models to obtain photorealistic and diverse results. MVDiffusion~\cite{MVDiffusion} and PanoFusion~\cite{PanFusion} finetune a latent diffusion model~\cite{LDM} to generate a 360-degree panorama of a scene. Text2Room~\cite{Text2room} and LucidDreamer~\cite{LucidDreamer} start with an initial RGB image and iteratively build the 3D scene by progressively warping and inpainting. CAT3D~\cite{CAT3D} and Bolt3D~\cite{Bolt3D} trained a multi-view LDM to generate novel views from input images, followed by a 3D reconstruction. Despite these advances, existing methods struggle to synthesize large viewpoint changes~\cite{ViewCrafter, Motionctrl} and semantically coherent scenes~\cite{CAT3D, Bolt3D, ZeroNVS} beyond observed areas. 

The line of work that most closely relates to ours employs 3D layout prior to guide the generation process. Set-the-Scene~\cite{SetTheScene}, SceneCraft~\cite{SceneCraft}, and Layout2Scene~\cite{Layout2Scene} generate 3D scenes by distilling the pretrained image diffusion models conditioned on a given semantic layout. These methods achieve better view and semantic consistency, but the realism and controllability remain limited. While Ctrl-Room~\cite{Ctrl-Room} and ControlRoom3D~\cite{ControlRoom3D} generate panoramas with high visual fidelity, they struggle to extrapolate the scene beyond a fixed camera location without resorting to a dedicated room completion procedure. We argue that these limitations stem from the scarce scale and diversity of available 3D scene datasets, hindering the learning of robust 3D priors.

\paragraph{Indoor Scene Dataset} Existing indoor datasets are either captured from real-world scenes using RGB~\cite{StereoMagnification, DL3DV} or RGB-D~\cite{SUNRGBD, ScanNet, ScanNetpp, Matterport3D} sensors or professionally designed with curated 3D CAD furniture models~\cite{3D-FRONT, Structured3D, Replica, Hypersim}. The real-world dataset provides a physically realistic appearance observation of 3D scenes; however, collecting and annotating these data typically requires significant resources in terms of cost and labor. On the other hand, indoor synthetic datasets address the constraints of real-world data by supplying extensive, varied, and richly annotated scenes. In addition, Structured3D~\cite{Structured3D} and Hypersim~\cite{Hypersim} utilize the advanced render engine for photorealistic image rendering with accurate 2D labels. However, the camera view is limited, which restricts the downstream application.

\section{\method Dataset}
\label{sec:dataset}

\begin{figure}[t]
    \centering
    \includegraphics[width=0.9\linewidth]{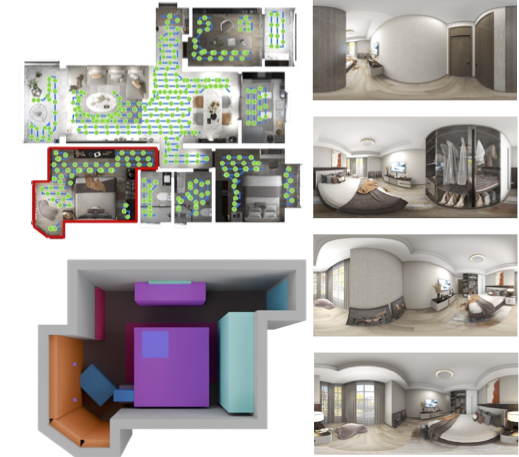}
    \caption{Illustration of our dataset. For each scene, we provide comprehensive panoramic renderings and 3D layout annotation.}
    \label{fig:dataset}
\end{figure}

We summarize the commonly used indoor scene dataset for layout-conditioned scene synthesis in \cref{tab:dataset}. As one can see, the real-world datasets suffer from a limited number of scenes, incomplete 3D annotations, and inconsistent annotation quality. Synthetic datasets are easier to annotate with ground-truth 3D labels, but they still have limitations in scene diversity (for example, Hypersim only has 461 scenes) or camera viewpoints (for example, Structured3D provides a single panorama for each room). 

In this paper, we build a new dataset to train generative models for 3D indoor scenes. Our dataset is based on a large repository of house designs sourced from an online platform in the interior design industry. Most of these designs are designed by professional designers and are intended for real-world production.

As shown in \cref{fig:dataset}, we create physically plausible camera trajectories that navigate smoothly through each scene while avoiding obstacles. These trajectories are sampled at $0.5\textrm{m}$ intervals to ensure comprehensive spatial coverage. For each viewpoint, we generate photorealistic panoramic renderings using an industry-leading rendering engine, capturing color, depth, normal, semantic, and instance segmentation data. We further convert the panoramic image into multiple perspective images using equilib~\cite{equilib}. To ensure both quality and diversity, we apply rigorous filtering criteria during dataset curation, resulting in \scene_number distinct scenes encompassing \room_number individual rooms with diverse room types. Each scene is annotated with precise 3D layouts and divided into \train_scene_number/\test_scene_number scenes for training/testing, respectively. 

\cref{fig:dataset} also shows some panoramas and 3D layout annotation from our dataset. Our dataset offers comprehensive structural layout annotations, including architecture elements (\ie, walls, doors, and windows). We further simulate diverse camera motion patterns from panoramic video data to train and evaluate the generation capabilities of existing approaches -- a crucial advantage over limited rule-based trajectories from existing dataset. The empirical benefits of this dataset are demonstrated in the experiments.
\section{Method}
\label{sec:method}

\begin{figure*}[t]
    \centering
    \includegraphics[width=\linewidth]{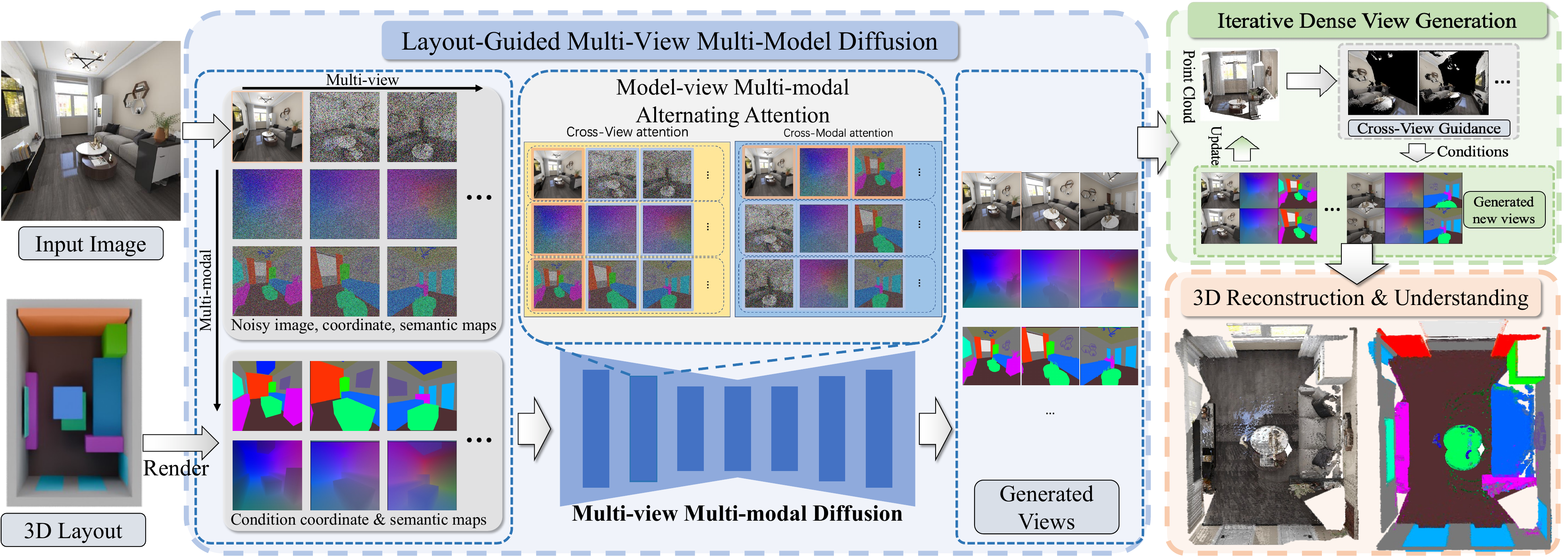}
    \vspace{-6mm}
    \caption{Overall pipeline. \method takes as input a 3D semantic layout and one or more posed images, to create a 3D scene. First, we generate per-view RGB images, scene coordinate maps, and semantic segmentation maps from a Layout-Guided Multi-view Multi-modal diffusion model. Then, we adopt an iterative dense view generation strategy to generate images at more sampled viewpoints. Finally, these images are fed into a 3D reconstruction method to produce the final result.}
    \label{fig:pipeline}
    \vspace{-4mm}
\end{figure*}

We introduce \method, a novel method that generates Gaussian Splatting~\cite{kerbl20233d} scenes with semantics conditioned on a 3D layout with reference images. Specifically, given a semantic layout and one or more source images, \method first utilizes a layout-guided multi-view multi-modal diffusion model to generate dense views of the target scene via Iterative Dense View Generation (detailed in \cref{sec:iterative_generation}). Then, we recover those dense views to a unified semantic Gaussian Splatting via an off-the-shelf reconstruction method~\cite{RaDe-GS}.

We start by providing a brief overview of multi-view diffusion models in \cref{sec:prelimiary}. Then, we introduce our layout-guided latent diffusion model in \cref{sec:multiview_multimodal_diffusion}, followed by the iterative generation scheme in \cref{sec:iterative_generation} and the 3D reconstruction process in \cref{sec:gs_reconstruction}. 

\subsection{Preliminaries}
\label{sec:prelimiary}

\paragraph{Multi-view Diffusion Model} A multi-view latent diffusion model takes a single or multiple posed source views as input and generates multiple novel images in some target camera views. 
To incorporate multi-view conditioning, it typically involves two designs:  (1) 2D attention layers are improved to a 3D-aware or multi-view aware attention mechanism, such as epipolar constraint~\cite{hartley2003multiple}, to capture multi-view features across different source views. (2) Camera poses are encoded by Plucker coordinate maps~\cite{plucker1865xvii, RayDiffusion} and then processed by a Transformer to compute view-conditioned embeddings. 

Given $M$ input views $\mathbf{I}_M = \{I_1, \ldots, I_M\}$ with camera poses $\mathbf{C}_M = \{C_1, \ldots, C_M\}$. Multi-view diffusion aims to predict $N$ new view images $\mathbf{I}_N = \{I_{M+1}, \ldots, I_{M+N}\}$ in camera poses $\mathbf{C}_N = \{C_{M+1}, \ldots, C_{M+N}\}$. In other words, the multi-view latent diffusion model aims to learn the following joint distribution, 
\begin{equation}
    p(\I_N \mid \I_M, \C_{M+N}),
    \label{eq:mvd}
\end{equation}
where $\C_{M+N} = \{C_1, \ldots, C_{M+N}\}$ includes camera poses for both input and output views.

\paragraph{Layout Condition in MVD} A 3D semantic layout provides an informative description of the scene. Following previous works~\cite{Ctrl-Room, ControlRoom3D, SceneCraft}, we represent the layout as a set of semantic bounding boxes of objects $\{\mathbf{b}_k \}^{K}_{k=1}$, where each box $\mathbf{b}_k$ includes the center location $l_k \in \mathbb R^3$, the size $s_k \in \mathbb R^3$, the orientation $r_k \in \mathbb R$ around the vertical axis, and the category label $z_k$. For each viewpoint with camera parameter $C_n = (K_n, T_n)$, we render a semantic map $S_n^{\rm layout}$ and a depth map $D_n^{\rm layout}$ of the bounding box of the 3D layout. The $D_n^{\rm layout}$ is then converted to the scene coordinate map $P_n^{\rm layout} = T_{n} \cdot (K_{n}^{-1} \cdot D_n^{\rm layout})$. In this way, we obtain the input layout conditions $\S_{M+N}^{\rm layout}, \P_{M+N}^{\rm layout} = \{ S_n^{\rm layout}, P_n^{\rm layout}\}_{n=1}^{M+N}$ for the latent diffusion model. We use scene coordinate maps instead of depth maps to represent 3D scene geometry because they encode the scene in a globally consistent manner. As discussed in previous work~\cite{Bolt3D, WorldVD}, they facilitate learning multi-view geometric consistency in the latent diffusion model. Therefore, extending \cref{eq:mvd}, the joint distribution to be learned for layout-conditioned multi-view image generation can be formulated as follows,
\begin{equation}
    p(\I_N \mid \I_M, \S_{M+N}^{\rm layout}, \P_{M+N}^{\rm layout}, \C_{M+N}).
\end{equation}
Note that the layout conditions $S_n$ and $P_n$ only provide a coarse description of the bounding box without pixel-level details, as shown in Figure~\ref{fig:pipeline}.

\subsection{Layout-guided Multi-view Diffusion Model}
\label{sec:multiview_multimodal_diffusion}

Since the input layout maps do not contain pixel-level details, we further predict a pixel-wise semantic map $S_n$, scene coordinate map $P_n$ for each viewpoint. 
This joint learning scheme improves 3D consistency in two ways:
\begin{itemize}
    \item {\bf Explicit 3D supervision.} By explicitly integrating both geometric and semantic maps into the latent diffusion model, \method leverages direct 3D supervision to achieve high-fidelity novel view synthesis results while maintaining cross-view consistency.
    
    \item {\bf Cross-view guidance.}  With the additional pixel-wise scene coordinate maps, we provide fine-grained guidance for diffusion by computing a warped image at any target view from the input images. Specifically, we adopt the point cloud based render~\cite{Pytorch3D} to obtain the warped image $I_n^{\rm warp}, \forall n \in \{M+1, \ldots, M+N\}$. The warped image is encoded and concatenated with the original noise map $I_n$ to form a conditioning signal for the target view, which we denote as augmented target views $\hat{I}_n = [I_n; I_n^{\rm warp}]$.
\end{itemize}

The joint distribution to be learned now becomes,
\begin{equation}
    p(\hat{\I}_N, \S_{M+N}, \P_{M+N} \mid \I_M, \S_{M+N}^{\rm layout}, \P_{M+N}^{\rm layout}, \C_{M+N}).
\label{eq:mvd-layout}
\end{equation}

\begin{figure}
    \centering
    \includegraphics[width=\linewidth]{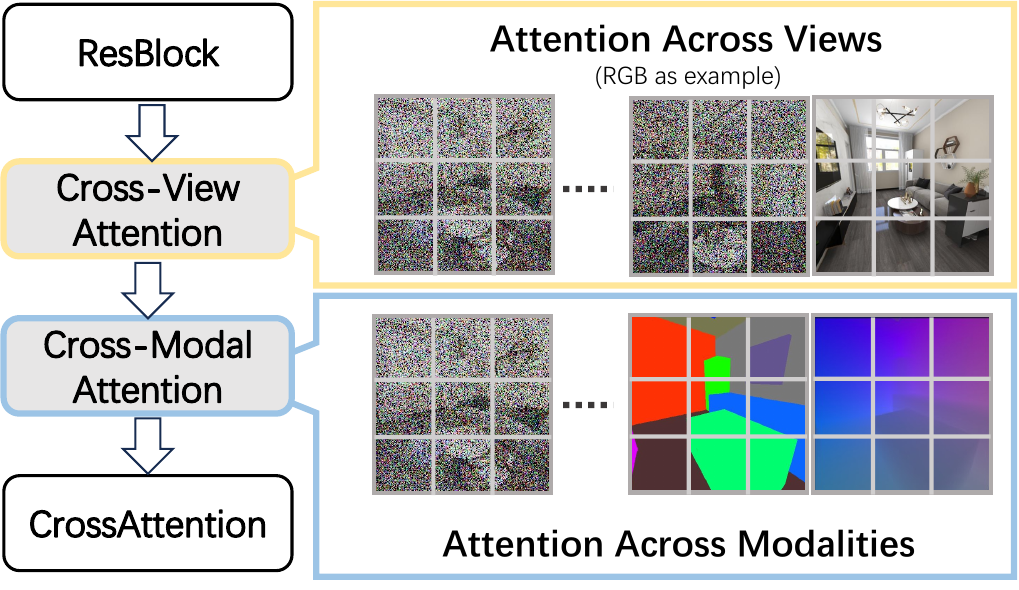}
    \vspace{-6mm}
    \caption{Multi-view and multi-modal alternating attention. It alternates between enforcing multi-view consistency and multi-modal fidelity within a unified attention mechanism.}
    \label{fig:layout_guide_attn}
\end{figure}

We use a v-parametrization and a v-prediction loss for the diffusion model~\cite{salimans2022progressive}. Following CAT3D~\cite{CAT3D}, the model is trained on a total of 8 views, with randomly sampled $\{1,3,7\}$ views as source views. We use the ground-truth scene coordinate map for warping in training, and use the predicted scene coordinates during inference. 

\paragraph{Multi-view Multi-modal Alternating Attention} With our formulation \cref{eq:mvd-layout}, our objective is to generate output that is consistent with multiple viewpoints and modalities. We observe that a simple modification to the standard architecture, namely an alternating attention mechanism, is effective in preserving the desired consistency.
% We introduce an alternating attention mechanism to improve the effectiveness of MVD. As illustrated in \cref{fig:layout_guide_attn}, the new architecture operates through complementary attention pathways: cross-view attention and cross-modal attention. Cross-view attention~\cite{Wonder3D, CAT3D} processes reshaped tokens along the view dimension (\eg, $\{\mathbf{t}_1^I, \mathbf{t}_2^I, \ldots, \mathbf{t}_{M+N}^I\}$ for all RGB images), allowing for feature aggregation across multiple views in each modality. Cross-modal attention operates within each view, observing modality-specific tokens (\eg, $\{\mathbf{t}_n^I, \mathbf{t}_n^S, \mathbf{t}_n^P\}$ for image, semantics, and geometry) to achieve fine-grained feature alignment. This design achieves a balance between integrating information across different views and different modalities with a low computational cost.% However, a naive implementation with the standard Transformer blocks in existing MVD models would lead to heavy computational loads. Instead, we observe that a simple modification to the standard architecture, namely an alternating attention mechanism, is effective in reducing the computation while preserving the desired consistency. 
As illustrated in \cref{fig:layout_guide_attn}, the new architecture operates through complementary attention pathways: cross-view attention and cross-modal attention. Inspired by previous works~\cite{Wonder3D, CAT3D}, our cross-view attention processes reshaped tokens along the view dimension (\eg, $\{\mathbf{t}_1^I, \mathbf{t}_2^I, \ldots, \mathbf{t}_{M+N}^I\}$ for all RGB images), allowing feature aggregation across multiple views in each modality. While cross-modal attention operates within each view, observing modality-specific tokens (\eg, $\{\mathbf{t}_n^I, \mathbf{t}_n^S, \mathbf{t}_n^P\}$ for image, semantics, and geometry) to achieve fine-grained feature alignment. This design achieves a balance between integrating information across different views and different modalities.
% $\{\mathbf{t}_n^I, \mathbf{t}_n^S, \mathbf{t}_n^P, \mathbf{t}_n^{S^{\rm layout}}, \mathbf{t}_n^{P^{\rm layout}}\}$

\begin{figure}[t]
    \centering
    \includegraphics[width=\linewidth]{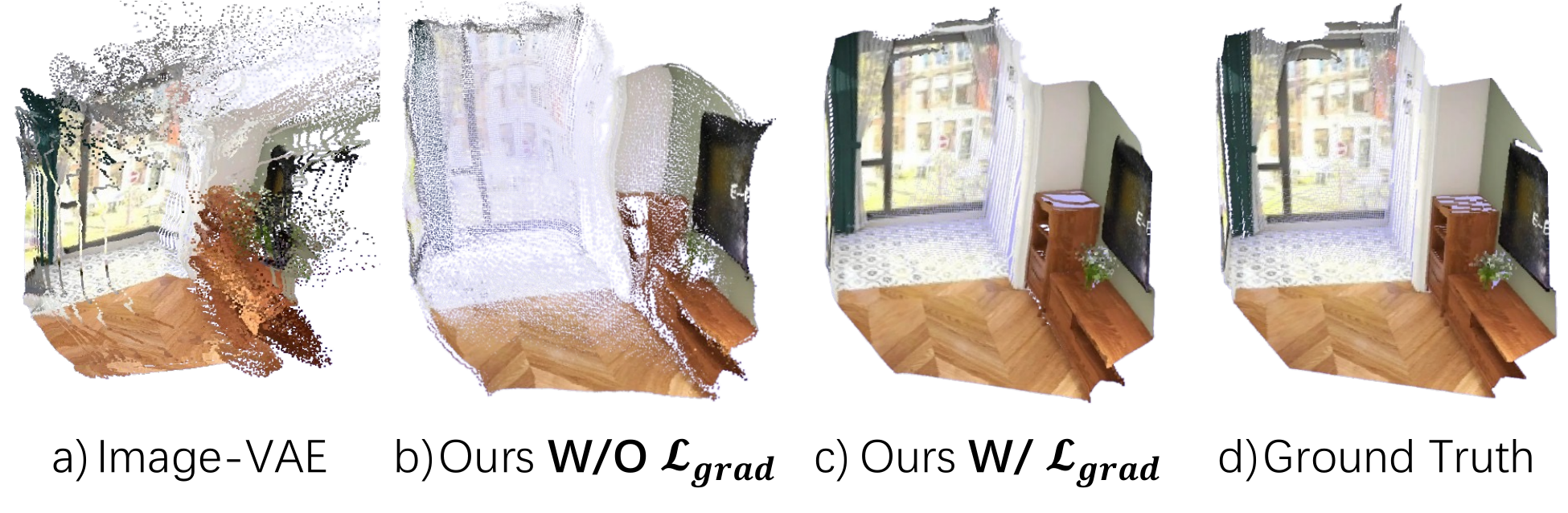}
    \vspace{-6mm}
    \caption{Comparison of reconstruction results for scene coordinate map. The image VAE (a) generates noisy results, and the SCM-VAE without gradient loss (b) produces distorted results. Our SCM-VAE (c) accurately reconstructs the scene geometry.}
    \label{fig:scm_vae}
    \vspace{-4mm}
\end{figure}

\paragraph{Scene Coordinate Map VAE (SCM-VAE)} A standard image VAE pretrained on RGB images generalizes well to semantic maps, but fails to accurately reconstruct scene coordinate maps, leading to poor geometric fidelity. See \cref{fig:scm_vae}(a) for an example. To address this, we introduce SCM-VAE, which encodes a scene coordinate map $P$ into a latent representation $z$ as $\mathbf z = \mathcal{\xi}(P)$ and reconstructs $z$ into a scene coordinate map with an uncertainty map as $\quad \{\hat{P}, \mathbf c\} = \mathcal{D}(\mathbf z)$, where $\mathcal{\xi}$ denotes the encoder and $\mathcal{D}$ is the decoder. 
The SCM-VAE is trained by fine-tuning the decoder $\mathcal{D}$ with an additional output dimension $\mathbf{c}$ from an image diffusion VAE, while keeping the encoder $\mathcal{\xi}$ frozen. $\mathbf{c}$  is activated by $\mathbf{c} = 1 + \rm exp(\mathbf{c})$ to ensure a strictly positive confidence~\cite{wan2018confnet}. The training objective combines standard VAE reconstruction with geometry-specific loss:
\begin{align}
    \mathcal{L} &= \mathcal{L}_{\text{rec}} + \lambda_1 \mathcal{L}_{\text{grad}}, \\
    \mathcal{L}_{\text{rec}} &=  c\odot\| \hat{P} - P \| - \alpha \log c, \\
    \mathcal{L}_{\text{grad}} &= \sum_{s=1}^{4} \| ({\nabla} \hat{P}_i^s - {\nabla}  P^s_i) \| ,
\label{eq:append_layout_diffusion_physical_objective}
\end{align}
where $\alpha=0.2$ and $\odot$ denotes element-wise multiplication. Here, we follow previous monocular depth estimation works~\cite{godard2017unsupervised} to use a multiscale gradient loss $\mathcal{L}_\textrm{grad}$ to improve boundary sharpness in the decoded scene coordinate map. As we can see in \cref{fig:scm_vae}, our SCM-VAE with $\mathcal{L}_\textrm{grad}$ outperforms the one without the term, especially around complex object boundaries and flat areas.

\subsection{Iterative Dense View Generation}
\label{sec:iterative_generation}

Our goal is to generate a complete 3D scene aligned to the given layout and text or image prompt. Although our layout-guided MVD model can generate an arbitrary number of views in principle, it is limited by GPU memory constraints. Thus, instead of generating all views at once, we adopt an iterative view synthesis strategy, a similar approach is also used in previous work~\cite{ViewCrafter, See3D}. The main idea is to incrementally maintain a colored global point cloud of the scene to enforce appearance consistency between iterations. During each iteration, the point cloud $\mathcal{P}$ is projected onto the target views $\I^{warp}$ to provide pixel-aligned guidance for consistent generation. The SCMs of the target view generated by our diffusion model will be inserted into $\mathcal{P}$. In this way, we can effectively reduce error accumulation. Furthermore, by incorporating the uncertainty map $\mathbf{c}$, we filter out 3D points with uncertainty below a predefined threshold, resulting in cleaner warped images.  

The iterative process, detailed in \cref{algo:iterative_generation}, proceeds as follows: \emph{First}, an initial point cloud is built by obtaining the scene coordinate maps $\P_M$ for the input views $\I_M$. \emph{Second}, at the beginning of each iteration, we render warped images for the target views from the point cloud. Then, we perform inference with not only the input images $\I_m$, but also the warped images to ensure global consistency. We then update the point cloud accordingly. \emph{Finally}, we collect the images generated from all iterations as output. 

\begin{algorithm}[t]
\caption{Iterative Dense View Generation}
\footnotesize
\SetAlgoNoLine
\KwIn{input views $\I_M$, camera poses for all iterations $\{\C_M, \C_{N_1}, \ldots, \C_{N_K}\}$, global point cloud $\mathcal{P}$ initialized as empty, layout-guided multi-view diffusion model $\mathcal{U}(\cdot)$.
}
\textbf{Initialization}: \\
Obtain the $\S_M$ and $\P_M$: $\{\S_{M}, \P_M\} \leftarrow \mathcal{U}(\I_M, \C_M)$; \\
Update global point cloud: $\mathcal{P} \leftarrow \P_M$;

\For{$k \gets 1$ \KwTo $K$}{
    Render warped images: $\I_{N_k}^{\rm warp} \leftarrow \mathrm{Render}(\mathcal{P}, \C_{N_k})$; \\
    Augment the target views $\hat{\I}_{N_k} = [\I_{N_k}; \I_{N_k}^{\rm warp}]$;\\
    Run inference: $\{\hat{\I}_{N_k}, \S_{N_k}, \P_{N_k}\}\leftarrow \mathcal{U}(\I_M, \C_{M+N_k})$; \\
    Update global point cloud: $\mathcal{P} \leftarrow \mathcal{P} \bigcup \P_{N_k}$;
}
\KwOut{all generated views: $\{\I_{N_1}, \I_{N_2}, \ldots, \I_{N_K}\}$.}
\label{algo:iterative_generation}
\end{algorithm}

\subsection{Scene Reconstruction and Understanding}
\label{sec:gs_reconstruction}

\begin{figure*}[t]
    \vspace{-7mm}
    \centering
	\subfloat{
		\includegraphics[width=\linewidth]{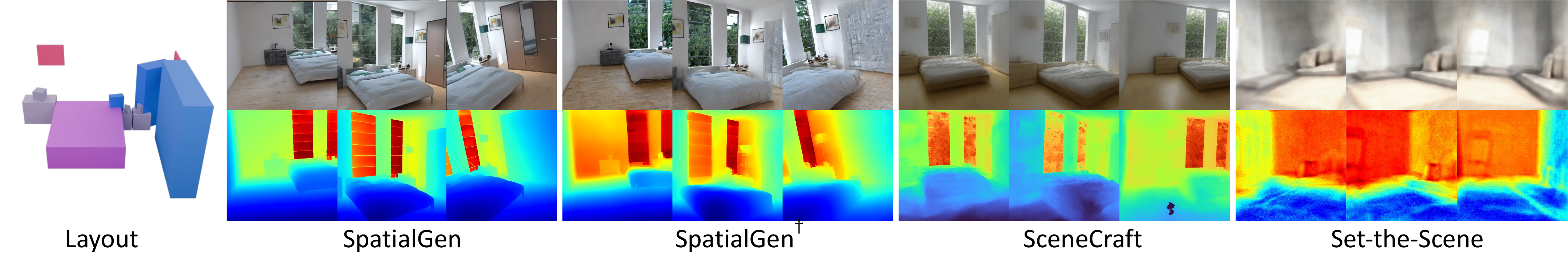}
		\label{fig:comparison:t2s_sds_hypersim}
        }\\
	\centering
	\subfloat{
		\includegraphics[width=\linewidth]{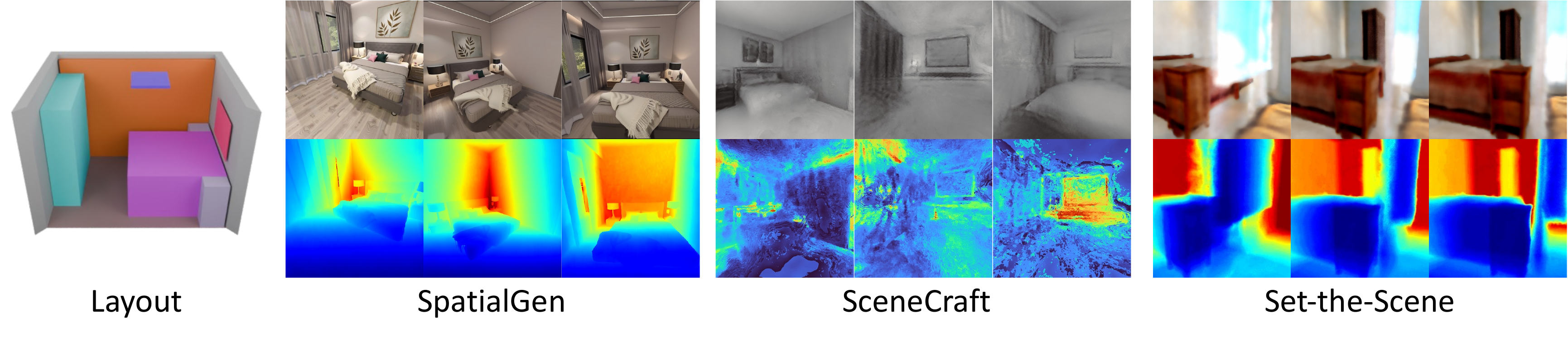}
            \vspace{-0.1in}
		\label{fig:comparison:t2s_sds_spatialgen}}
    \vspace{-2mm}
    \caption{Qualitative comparison to score distillation methods on Hypersim~\cite{Hypersim} (top row) and our dataset (bottom row), In each case, we show the generated color images and depth maps.}
    \label{fig:comparison:sds}
    \vspace{-5mm}
\end{figure*}

Building upon RaDe-GS~\cite{RaDe-GS}, we reconstruct a 3D scene representation from the densely generated color, geometric, and semantic images. Following Feature-3DGS~\cite{Feature-3dgs}, We augment the standard 3D Gaussians with a semantic feature in each point. The scene is initialized from the predicted point cloud $\mathcal{P}$.
% With our densely generated color, geometric, and semantic images, we reconstruct a Gaussian representation with semantic feature in each point like Feature-3DGS~\cite{Feature-3dgs} with RaDe-GS~\cite{RaDe-GS}. We initialize the scene from the predicted point cloud $\mathcal{P}$. 
During differentiable rendering optimization, we employ a depth supervision loss that utilizes the predicted scene coordinate maps, enabling rapid convergence in just 7,000 steps. As shown in \cref{fig:comparison:sds}, the pipeline produces high-fidelity RGB renderings and geometrically accurate depth reconstructions.
\section{Experiments}
\label{sec:exp}

% We first provide implementation details in \cref{sec:exp:detail}. \cref{sec:exp:text2scene} shows our scene generation results under various layout conditions and prompts, which are compared with publicly available text-to-3D methods. Next, we show image-to-3D scene results in \cref{sec:exp:image2scene}.

% \begin{figure*}[t]
%     \vspace{-6mm}
%     \centering
%     \includegraphics[width=\linewidth]{figures/quality/t2s_sds.pdf}
%     \vspace{-6mm}
%     \caption{Qualitative comparison to score distillation methods on Hypersim~\cite{Hypersim} (top row) and our dataset (bottom row), In each case, we show the generated color images and depth maps.}
%     \label{fig:comparison:sds}
%     \vspace{-3mm}
% \end{figure*}

\subsection{Experiment Setup}
\label{sec:exp:detail}

\paragraph{Benchmark Datasets} We use both existing datasets (\ie, Hypersim~\cite{Hypersim} and Structured3D~\cite{Structured3D}) and our new dataset. As discussed in~\cref{sec:dataset}, one key difference of these datasets is in the diversity of camera viewpoints. Since our dataset provides abundant panorama images at dense locations in each room, we can design various camera movement patterns to conduct an extensive evaluation.

Specifically, we introduce four distinct camera trajectories with different amounts of view overlap and distance between input and target views: (i) \textit{Forward}: the trajectory follows a linear path with minimal directional variation, simulating steady camera movement. (ii) \textit{Inward Orbit}: both the input and output views are directed toward the center of the room, ensuring substantial view overlap; (iii) \textit{Outward Orbit}: the input and output views are at the same location, but oriented differently, with less than $45^\circ$ overlap at adjacent views. (iv) \textit{Random Walk}: input and output views are sampled from a continuous random-walk path, resulting in minimal view overlap. Please refer to the appendix for the visualization of these settings.

\paragraph{Implementation Details} We implement \method in PyTorch. Both the SCM VAE and the latent diffusion model are fine-tuned from stable diffusion 2.1 (SD-2.1)~\cite{LDM}. We use AdamW optimizer~\cite{AdamW}. For SCM VAE, we freeze the encoder and only fine-tune the decoder for 10K steps with a batch size of 64. For the latent diffusion model, we fine-tune it for 35K steps with a batch size of 128. The first 16K steps use resolution $256 \times 256$, whereas the remaining steps use resolution $512 \times 512$. All models are fine-tuned on 64 NVIDIA RTX 4090 GPUs. The learning rate starts at ${10}^{-4}$ and decays by a factor of 0.01 at 90\% of the total training process. We render warped images using PyTorch3D~\cite{Pytorch3D}. 

For text-to-3D scene generation, we further train a layout ControlNet~\cite{ControlNet} to generate the reference image for our latent diffusion model. 

\begin{table}[t]
    \centering
    \footnotesize
    \renewcommand{\tabcolsep}{1pt}
    \caption{Comparison to score distillation methods.}
    \label{table:text2scene_sds}
    \vspace{-3mm}
    \begin{tabular}{lccc}
        \toprule
        Dataset & Method & {CLIP Sim. (\%)} $\uparrow$ & {Image Reward $\uparrow$} \tabularnewline
        \midrule
        \multirow{4}{*}{Hypersim~\cite{Hypersim}} 
        & Set-the-Scene~\cite{SetTheScene} & 25.18 & -2.005 \tabularnewline
        & SceneCraft~\cite{SceneCraft} & 26.94 & -1.096 \tabularnewline
        & \method{}$^\dagger$ & 25.93 & -1.168 \tabularnewline
        & \method & \textbf{27.59} & \textbf{-0.285} \tabularnewline
        \midrule
        \multirow{3}{*}{Our Dataset}
        & Set-the-Scene~\cite{SetTheScene} & 25.23 & -2.100 \tabularnewline
        & SceneCraft~\cite{SceneCraft} & 18.93 & -2.267 \tabularnewline
        & \method & \textbf{27.89} & \textbf{-0.123} \tabularnewline
        \bottomrule
    \end{tabular}
    \vspace{-6mm}
\end{table}

\subsection{Text-to-3D Scene Generation}
\label{sec:exp:text2scene}

As discussed before, existing layout-conditioned text-to-3D methods can be grouped into two categories: \emph{score distillation} methods~\cite{SetTheScene, SceneCraft, Layout2Scene, GALA3D} and \emph{panorama-as-proxy} methods~\cite{ControlRoom3D, Ctrl-Room}. In the following, we compare to these two groups separately.

\subsubsection{Comparison to Score Distillation Methods}

For this experiment, we compare to two open-source methods: Set-the-Scene~\cite{SetTheScene} and SceneCraft~\cite{SceneCraft}. The other methods such as Layout2Scene~\cite{Layout2Scene} and GALA3D~\cite{GALA3D} do not release their codes. We conduct experiments on both Hypersim~\cite{Hypersim} and our new datasets. Evaluation is performed on the test scenes of each target dataset following standard protocols. 

For \emph{Hypersim dataset}, we directly use the official checkpoints of Set-the-Scene and SceneCraft. To illustrate the benefit of our proposed large-scale dataset, we include two training configurations for our model: \method{}$^\dagger$ which is trained solely on Hypersim, and \method which is trained on a combination of Hypersim and our datasets.  

For \emph{our dataset}, we fine-tune the layout ControlNet of SceneCraft and adapt the layouts to match the input of Set-the-Scene. All methods on our dataset use the \textit{Inward Orbit} camera trajectory for consistency.

\paragraph{Quantitative Results} \cref{table:text2scene_sds} reports the performance of all methods with established 2D rendering metrics: CLIP similarity score~\cite{CLIP} to measure text-image alignment and Image Reward~\cite{ImageReward} to assess human aesthetic preference. 

As one can see, our method performs slightly worse than SceneCraft when trained solely on Hypersim. We hypothesize that Hypersim is too small for powerful latent multi-view diffusion models like the one employed by our method. When trained on both Hypersim and our datasets, our method outperforms both SDS methods on all metrics.  \method achieves a significantly higher image-reward score than models trained solely on Hypersim, validating the benefit of our large-scale dataset for high-quality 3D scene generation. Furthermore, when tested on our dataset, \method consistently outperforms the baselines, with a significantly higher image-reward score.

\paragraph{Qualitative Results} The advantage of our method can be better observed in \cref{fig:comparison:sds}. On both Hypersim and our datasets, \method generates photorealistic scenes with superior details that are well-aligned with the specified layout. In contrast, SceneCraft struggles to balance layout adherence with text-prompt fidelity, and Set-the-Scene takes nearly two hours to synthesize a radiance field that still lacks fine-grained details. While \method{}$^\dagger$ produces blurry and artifact images, further underscores the benefit of our proposed dataset to the final output quality.

\subsubsection{Comparison to Panorama-as-Proxy Methods}

\begin{table}[t]
    \centering
    \footnotesize
    \renewcommand{\tabcolsep}{1pt}
    \caption{Comparison to panorama-as-proxy method.}
    \label{table:text2scene_pano}
    \vspace{-3mm}
    \begin{tabular}{lccc}
        \toprule
        Dataset & Method & {CLIP Sim. (\%)} $\uparrow$ & {Image Reward $\uparrow$} \tabularnewline
        \midrule
        \multirow{2}{*}{Structured3D~\cite{Structured3D}}
        & Ctrl-Room~\cite{Ctrl-Room} & \textbf{25.03} & \textbf{-1.016} \tabularnewline
        & \method & 23.90 & -1.405 \tabularnewline
        \midrule
        \multirow{2}{*}{Our Dataset}
        & Ctrl-Room~\cite{Ctrl-Room} & 22.63 & -1.546 \tabularnewline
        & \method & \textbf{27.89} & \textbf{-0.123}  \tabularnewline
        \bottomrule
    \end{tabular}
    \vspace{-3mm}
\end{table}

\begin{figure}
    \centering
    \includegraphics[width=\linewidth]{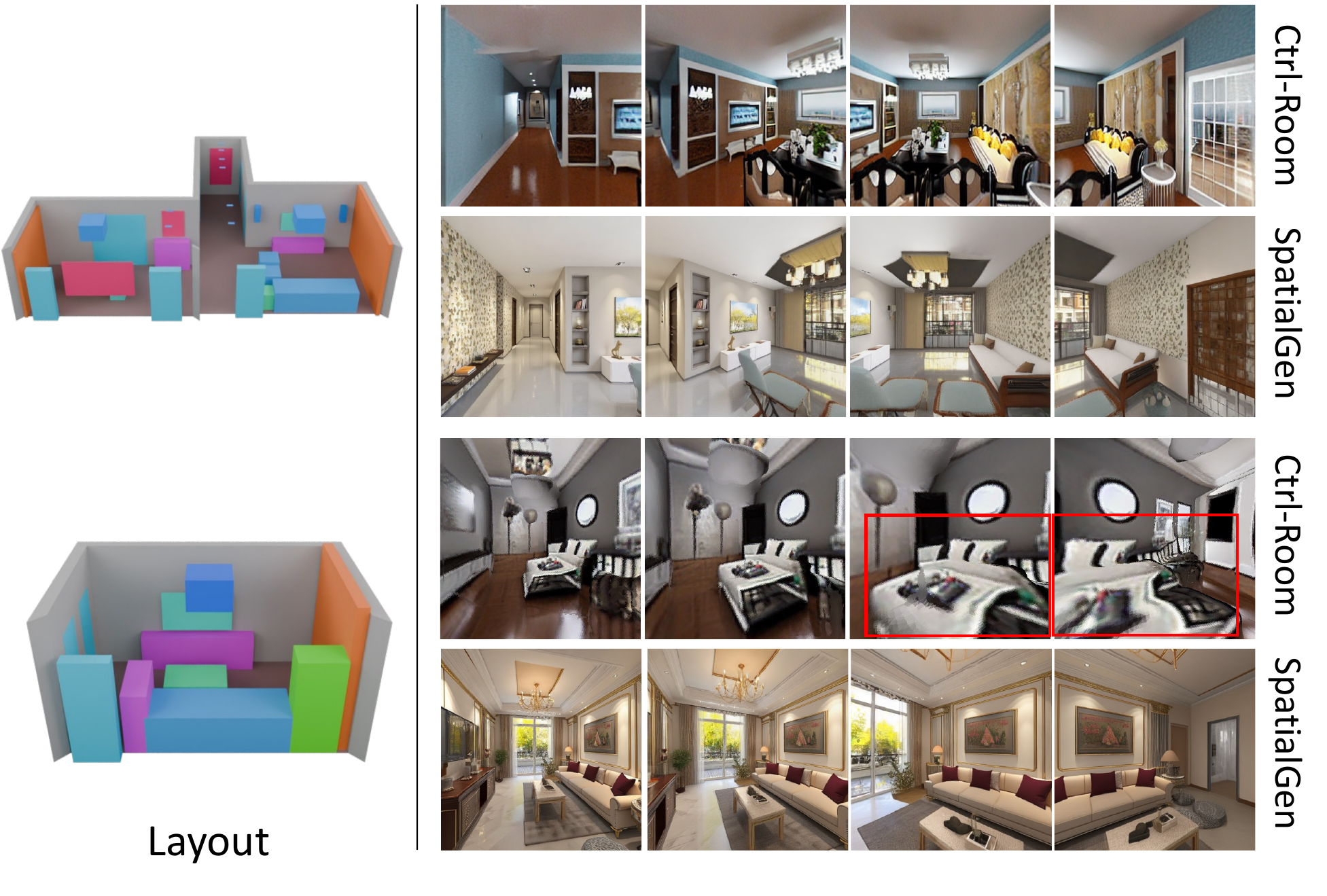}
    \vspace{-7mm}
    \caption{Qualitative comparison between our method and the panorama-as-proxy baseline~\cite{Ctrl-Room} on Structured3D~\cite{Structured3D} (top row) and our dataset (bottom row).}
    \vspace{-2mm}
    \label{fig:comparison:pano_as_proxy}
\end{figure}

Next, we compare our method to panorama-as-proxy methods. We choose Ctrl-Room~\cite{Ctrl-Room} as the baseline because the source code and checkpoints are available. We conduct experiments on both Structured3D and our datasets. For \emph{Structured3D dataset}, we use ground-truth layouts to ensure consistent inputs for both methods. For \emph{our dataset}, we use the \textit{Inward Orbit} camera trajectory for consistency.

\paragraph{Quantitative Results} \cref{table:text2scene_pano} reports the performance of both methods, we take the renderings of the generated mesh from Ctrl-Room for comparison. On Structured3D dataset (which provides only a single panorama per scene), our method still achieves competitive performance. The relatively lower scores are expected as Ctrl-Room is specifically trained to synthesize a single panorama at a fixed camera location. In contrast, our method generates multiple perspective images without explicitly exploiting the fact that all images are taken from a single location.

Nevertheless, the critical advantage of our method is revealed on our dataset: when rendering from novel viewpoints, the performance of Ctrl-Room degrades significantly, whereas our method consistently produces high-quality results from arbitrary views.

\paragraph{Qualitative Results} \cref{fig:comparison:pano_as_proxy} further highlights the advantage our method over panorama-as-proxy method. On our dataset, Ctrl-Room fails to synthesize coherent novel views, exhibiting severe distortions and artifacts. In contrast, our method is not limited to a single camera position (\ie, panorama generation). It achieves high-quality panorama generation on Structured3D while also enabling photorealistic novel view synthesis on our dataset. 

\subsection{More Experiments and Ablations}
Additionally, we evaluate \method on image-to-3D scene generation to validate its generative capability and 3D consistency. Recognizing that well-defined 3D semantic layouts are often unavailable in practice, we further extend the evaluation to video inputs. In this setting, we leverage the state-of-the-art layout estimation model~\cite{SpatialLM} to parse a 3D layout directly from the video. Additional results, implementation details, and ablations are provided in Appendix.B and Appendix.C.

\section{Conclusion \& Limitations}

We present \method, a novel framework for layout-guided 3D indoor scene synthesis. At the core of our pipeline is a multi-view multi-modal diffusion model, which generates images with high visual quality and geometric consistency. To train this model, we collect a new synthetic indoor scene dataset with \image_number panoramic renderings of \room_number rooms and the 3D layout annotations. These advancements open new possibilities for downstream applications such as interior design, embodied AI, and AR/VR.

\paragraph{Limitations} First, the cross-view and cross-modal attention introduces additional computational cost to the diffusion model, which limits \method to generate more images at a time. Second, the camera sampling strategy might affect the generation quality. Moreover, since our method relies on an initial RGB image, it is sensitive to any misalignment between the image and the provided 3D layout. We plan to address these challenges in the future.

\section*{Acknowledgments}

This work was supported in part by the Key R\&D Program of Zhejiang Province (2025C01001) and HKUST Project 24251090T019. We thank Yingqi Shen, Liangbin Hu, and Fuchun Dong (Manycore Tech) for dataset support; Chenfeng Hou for SDS-based experiments; Zhiwei Wang for ControlNet testing; and Kunming Luo for figure suggestions.

{
\small
\bibliographystyle{ieeenat_fullname}
\bibliography{spatialgen}
}

\clearpage
\appendix
\maketitlesupplementary

\begin{figure*}[!t]
    \centering
    \includegraphics[width=\linewidth]{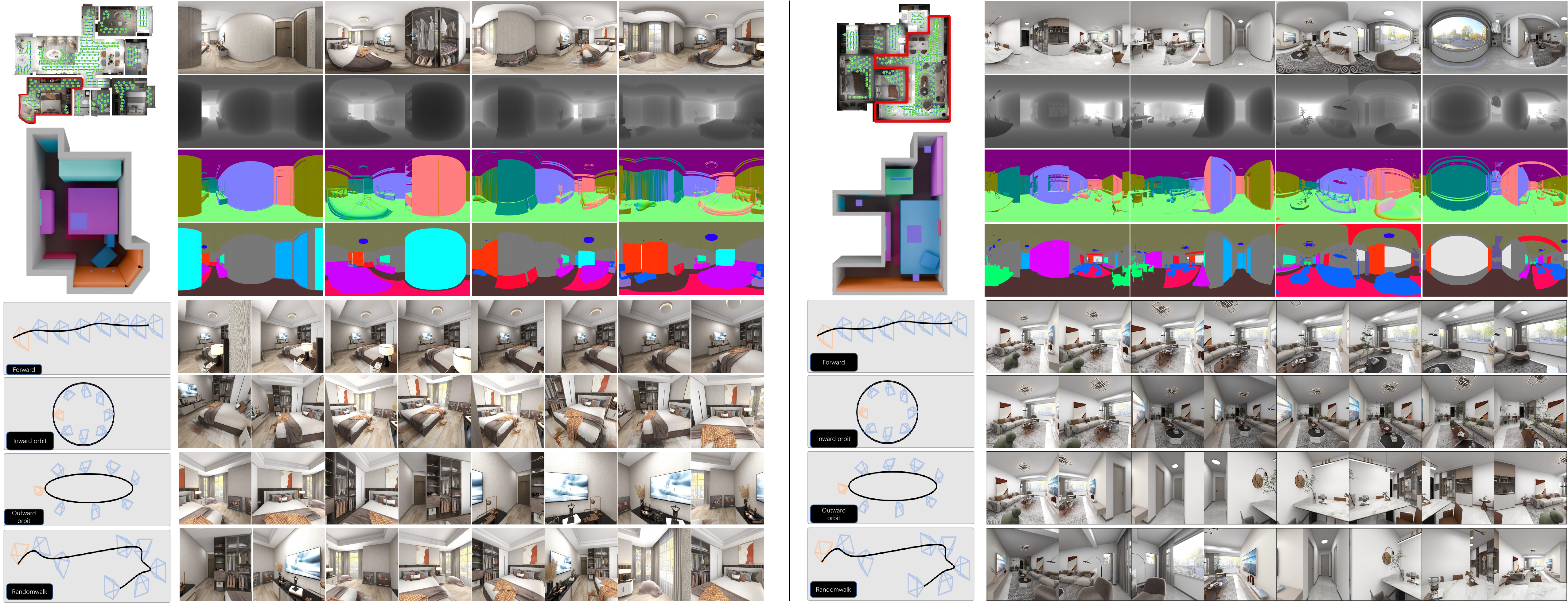}
    \caption{Examples of \method dataset.}
    \label{fig:supp:dataset_samples}
\end{figure*}

In the appendix, we provide more details of \method dataset in \cref{sec:supp:dataset}, additional experimental results in \cref{sec:supp:exp_and_result}, and ablation studies in \cref{sec:supp:ablation}.

\section{\method Dataset}
\label{sec:supp:dataset}

\subsection{Dataset Construction}

\paragraph{Data Curation} Our dataset is sourced from an online platform in the interior design industry, providing a large-scale collection of professional designs intended for real-world applications. We employ a rigorous multi-stage filtering pipeline to ensure both the quality and diversity of the dataset. 

The curation process begins by selecting scenes based on four key criteria: (i) professional designer ratings, (ii) the number of renderings generated by the design, (iii) a total floor area exceeding $20\textrm{m}^2$, and (iv) the presence of more than 35 unique objects.

Then, we extract individual rooms from each selected scene and apply additional filters to retain only those rooms that (i) have a floor area greater than $8\textrm{m}^2$ and (ii) contain more than 3 unique objects. 

\begin{figure}[t]
    \centering
    \footnotesize
    \renewcommand{\tabcolsep}{1pt}
    \begin{tabular}{ccc}
        \includegraphics[width=0.3\linewidth]{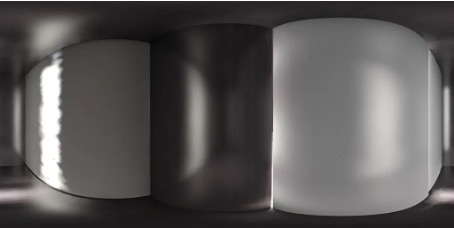} &
        \includegraphics[width=0.3\linewidth]{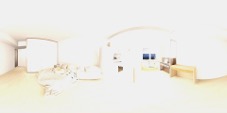} &
        \includegraphics[width=0.3\linewidth]{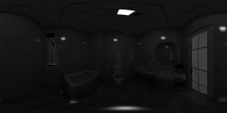} \tabularnewline
        camera–furniture collision & over-exposure & insufficient illumination
    \end{tabular}
    \caption{Example of low-quality renderings.}
    \label{fig:bad_samples}
\end{figure}

For rendering, we use an industry-leading rendering engine to generate images. We simulate physically plausible camera trajectories that navigate smoothly within each room while avoiding obstacles. After rendering, we implement strict quality control measures by discarding low-quality images—specifically those with camera-object intersections, overexposure, or inadequate lighting, as illustrated in \cref{fig:bad_samples}.

The final dataset consists of \scene_number distinct scenes, \room_number individual rooms covering a variety of room types, and \image_number photo-realistic panoramic renderings. The total floor area across all scenes is approximately $914,687\textrm{m}^2$.

\begin{figure}[t]
    \centering
    \includegraphics[width=\linewidth]{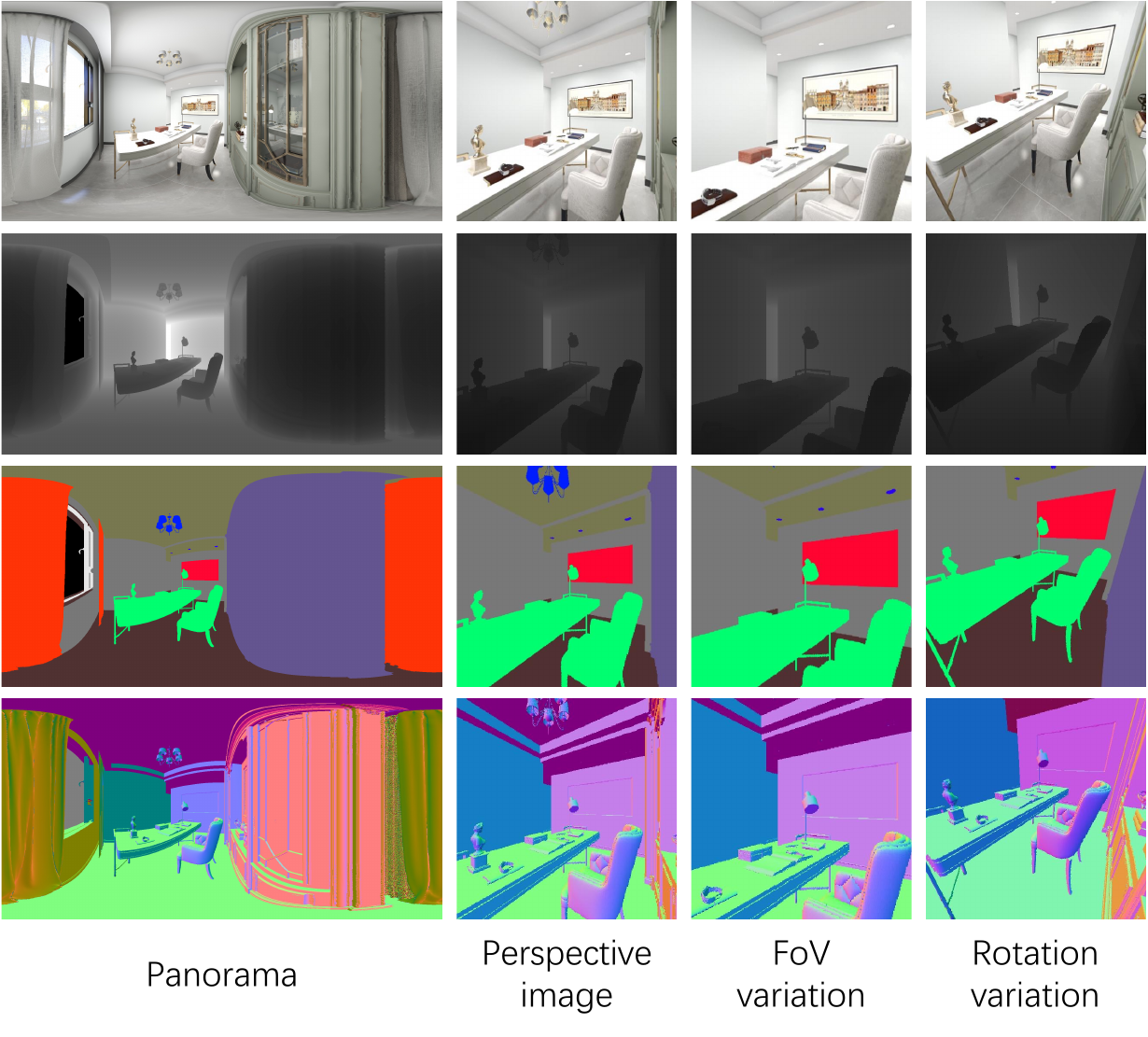}
    \caption{Camera configuration.}
    \label{fig:supp:camera_config}
\end{figure}

\paragraph{Camera configuration} We capture panoramic renderings at intervals of $0.5\textrm{m}$ to ensure comprehensive scene coverage, as shown in the top-left of \cref{fig:supp:dataset_samples}. Each panoramic rendering is generated at a resolution of $1024 \times 2048$ and includes color, albedo, depth, normal, semantic, and instance maps. The entire rendering process requires approximately 54K GPU hours.

Following an obstacle-avoiding camera trajectory within each room, we obtain dense sequences of panoramic images. Thanks to the $360^\circ$ field-of-view (FoV) of panoramas, we can simulate an unlimited number of perspective images with varying camera configurations. For each panoramic viewpoint, we generate perspective views with different fields-of-view and rotation angles using equirectangular-to-perspective projection~\cite{equilib}, as illustrated in \cref{fig:supp:camera_config}.

Furthermore, we introduce four distinct camera trajectories with varying amounts of view overlap and distances between input and target views: (i) \textit{Forward}: a linear path with minimal directional variation, simulating steady camera movement; (ii) \textit{Inward Orbit}: both input and output views are oriented toward the center of the room, ensuring significant view overlap; (iii) \textit{Outward Orbit}: the input and output views share the same location but have different orientations, resulting in less than $45^\circ$ overlap between adjacent views; and (iv) \textit{Random Walk}: input and output views are sampled along a continuous random-walk path, with minimal view overlap.

\subsection{Dataset Statistics}

\begin{figure}[t]
    \centering
    \includegraphics[width=\linewidth]{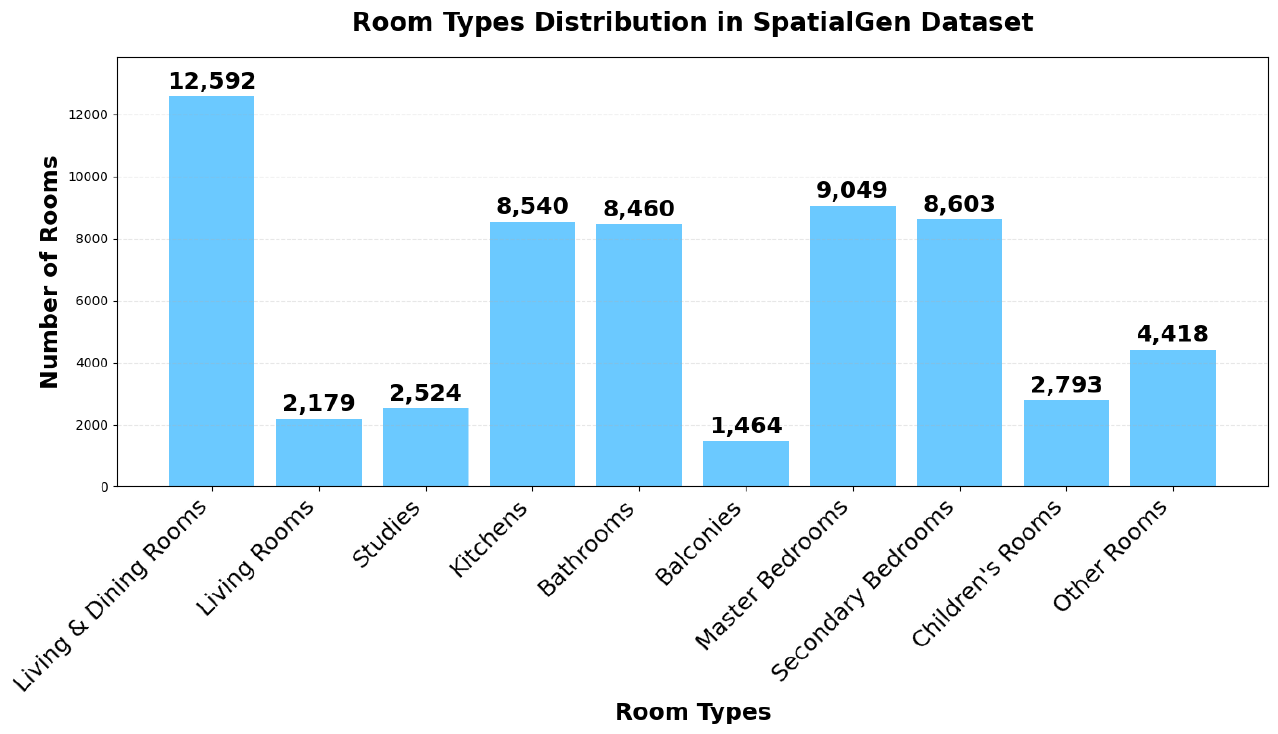}
    \caption{Room type distribution.}
    \label{fig:supp:room_dist}
\end{figure}

\paragraph{Room type statistics} The resulting dataset contains 12,592 living and dining rooms, 2,179 living rooms, 2,524 study rooms, 8,540 kitchens, 8,460 bathrooms, 1,464 balconies, 9,049 master bedrooms, 8,603 secondary bedrooms, 2,793 children's rooms, and 4,418 other room types, as illustrated in \cref{fig:supp:room_dist} representing a diverse and substantial collection of indoor environments.

\begin{figure}[t]
    \centering
    \includegraphics[width=\linewidth]{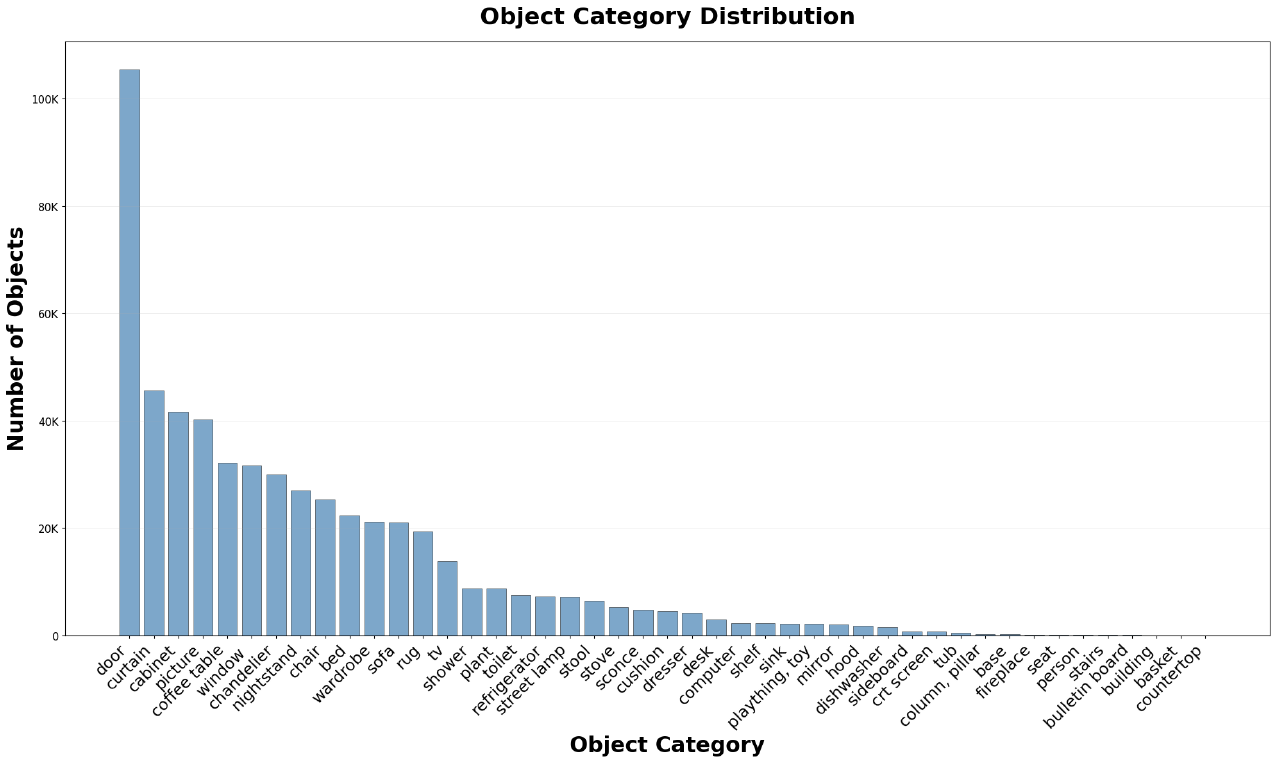}
    \caption{Object category distribution.}
    \label{fig:supp:object_dist}
\end{figure}

\paragraph{Object category statistics} The raw online designs initially contained approximately 65,000 object categories. We filtered out niche object classes specific to interior design and mapped the remaining objects to 62 common categories from ADE20K~\cite{ADE20K}. We then curated the object bounding boxes according to the following criteria: (i) objects outside the room layout were discarded; (ii) objects with any edge shorter than $0.1\textrm{m}$ or longer than $1.8\textrm{m}$ were excluded. This process yielded a total of 1,046,637 object bounding boxes. Figure~\ref{fig:supp:object_dist} shows the distribution of object categories throughout our dataset, excluding the spotlight and other categories (containing 250K and 240K instances, respectively) to improve visualization of the remaining categories.

\subsection{Dataset Visualization}

As shown in \cref{fig:supp:dataset_samples}, our dataset provides high-quality panoramic renderings accompanied by precise 2D annotations and comprehensive 3D structural layouts, including architecture elements (\eg, walls, windows, and doors), which distinguishes it from existing datasets like Hypersim~\cite{Hypersim}, offering extensive evaluation opportunities for scene generation and spatial understanding tasks.

\section{Additional Experiments and Results}
\label{sec:supp:exp_and_result}

In this section, we show more results of text-to-3D scene generation in~\cref{sec:supp:text2scene}, and conduct comprehensive experiments of image-to-3D scene generation in~\cref{sec:supp:img2scene}. Furthermore, we show generation results from common videos (unposed) captured in indoor scenes in~\cref{sec:supp:video2scene}.

\begin{figure*}[t]
    \centering
    \includegraphics[width=\linewidth]{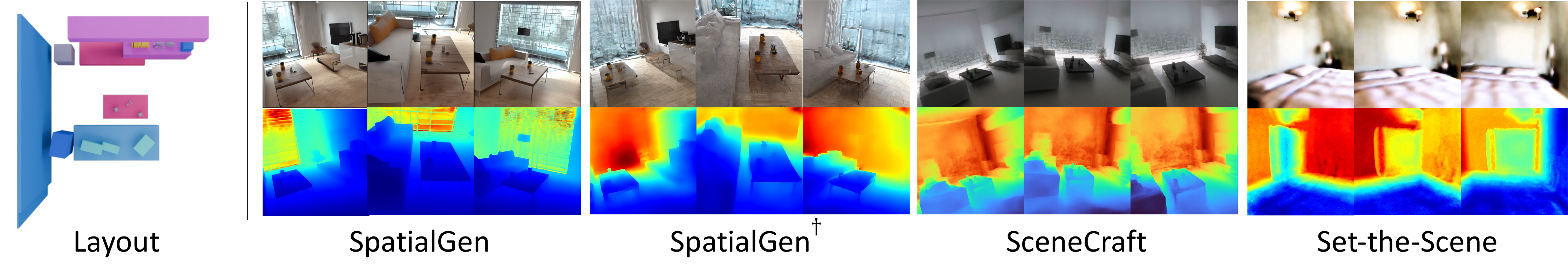}
    \caption{Qualitative comparison of text-to-3D scene on Hypersim~\cite{Hypersim} dataset. In each case, we show the generated color images and depth map.}
    \label{fig:supp:text2scene_hypersim}
\end{figure*}

\begin{figure*}[t]
    \centering
    \includegraphics[width=\linewidth]{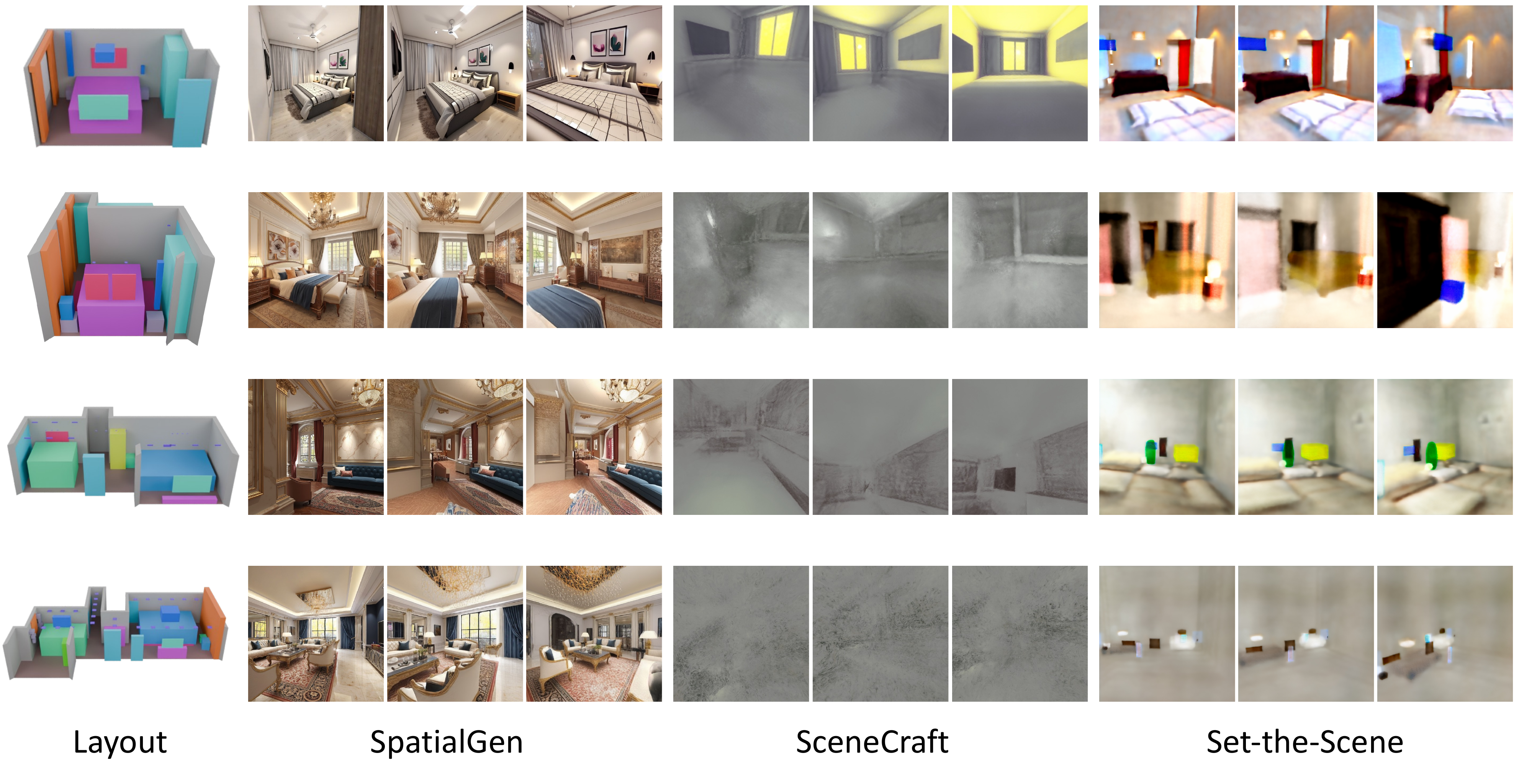}
    \caption{Qualitative comparison of text-to-3D scene on \method dataset. In each case, we show the generated color images.}
    \label{fig:supp:text2scene_spatialgen}
\end{figure*}

\begin{figure*}[t]
    \centering
    \includegraphics[width=\linewidth]{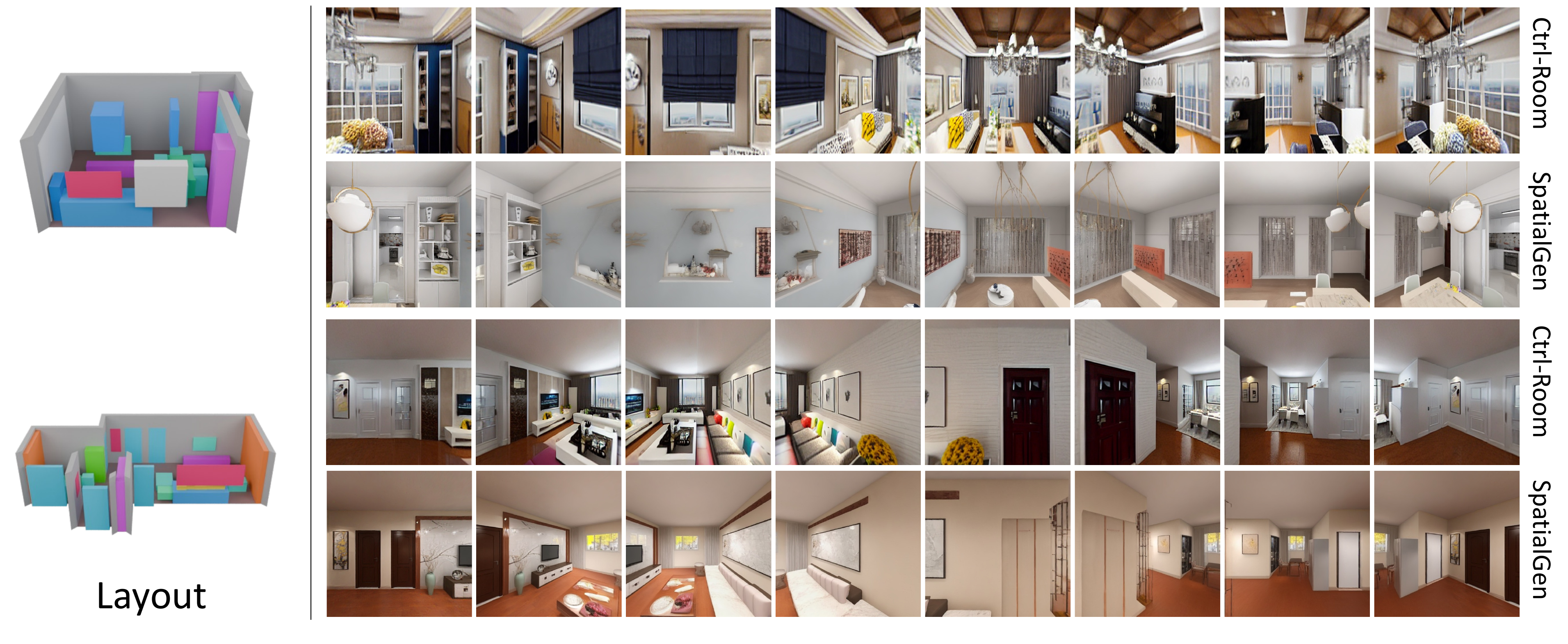}
    \caption{Qualitative comparison with Ctrl-Room on Structured3D for panorama generation. We split the panorama into eight perspective images for a direct comparison. Our method achieves competitive RGB synthesis compared with Ctrl-Room, resulting in photo-realistic scenes that are well-aligned with the provided layout.}
    \label{fig:supp:comp_ctrlroom_st3d}
\end{figure*}

\begin{figure*}
    \centering
    \includegraphics[width=\linewidth]{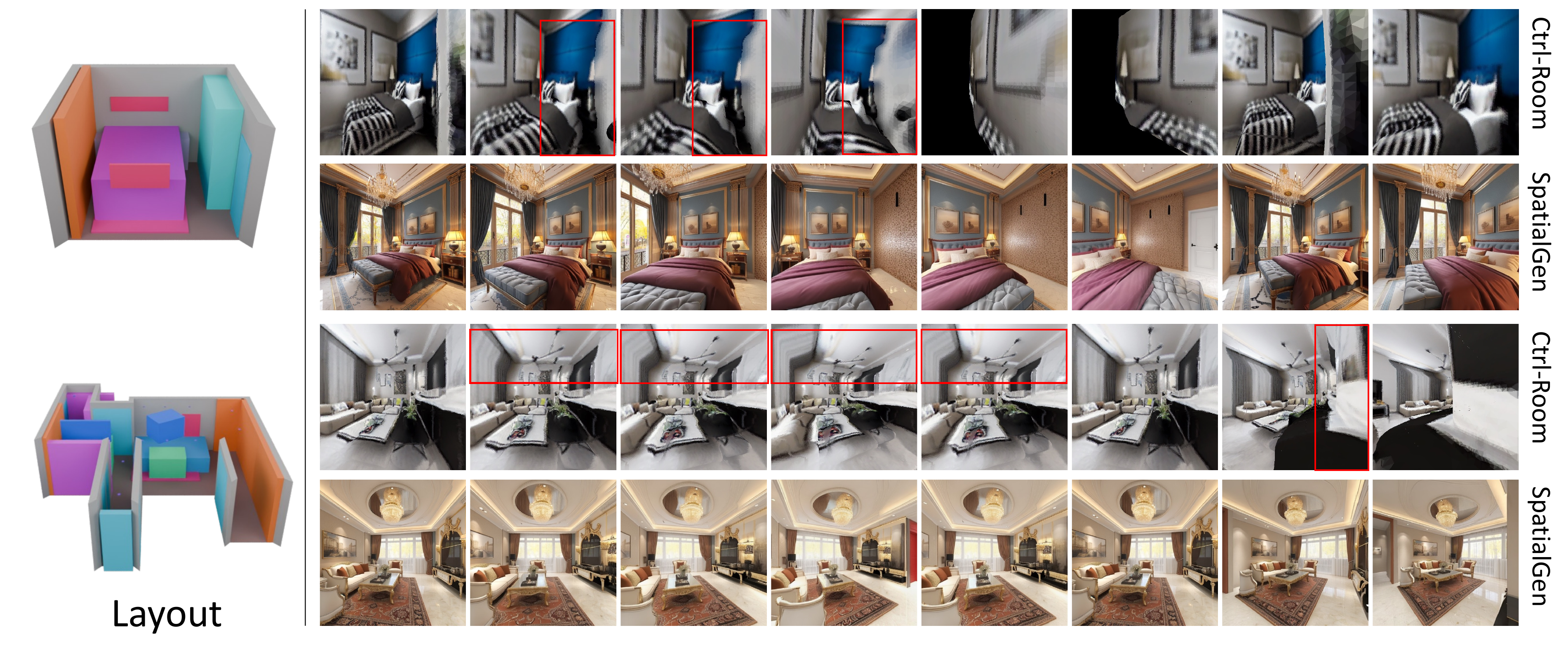}
    \caption{Qualitative comparison with Ctrl-Room on \method dataset. Ctrl-Room exhibits severe stretching artifacts and scale misalignments at novel viewpoints. In contrast, our method consistently produces photorealistic and fully 3D-consistent renderings from all views.}
    \label{fig:supp:comp_ctrlroom_spatialgen}
\end{figure*}

\subsection{Text to 3D Scene Generation} 
\label{sec:supp:text2scene}
As demonstrated in \cref{fig:supp:text2scene_hypersim}, our method outperforms SDS-based baselines on the Hypersim dataset~\cite{Hypersim}, producing photorealistic and layout-faithful scenes with superior detail. This advantage is evident even against our model trained only on Hypersim (\method{}$^\dagger$), which produces blurring and ambiguous results, highlighting the benefits of our dataset. 

In \cref{fig:supp:text2scene_spatialgen}, we compare against SceneCraft~\cite{SceneCraft} and Set-The-Scene~\cite{SetTheScene} under diverse 3D layouts on \method dataset. As the increase of layout complexity,  while competing methods fail to generate meaningful radiance fields or capture scene details, our approach consistently delivers realistic and coherent results for complex scenes like living and dining rooms.

We further compare our method against the panorama-as-proxy based method, Ctrl-Room~\cite{Ctrl-Room}, on both the Structured3D~\cite{Structured3D} and \method dataset. We split the panoramic image into 8 perspective images for a direct comparison, as shown in \cref{fig:supp:comp_ctrlroom_st3d}. 

For the \method dataset, we render a layout-semantic panorama from a random viewpoint to use as input for Ctrl-Room. We then spatially align its resulting mesh with our generated scene for a fair comparison. The results, presented in \cref{fig:supp:comp_ctrlroom_spatialgen}, demonstrate that Ctrl-Room exhibits severe stretching artifacts and scale misalignment at novel viewpoints. In contrast, our method consistently produces photorealistic and fully 3D-consistent renderings from all views.

\begin{table}[t]
    \centering
    \footnotesize
    \renewcommand{\tabcolsep}{3pt}
    \caption{Experimental results on image-to-3D scene generation under four distinct camera trajectories: \textit{Forward}, \textit{Inward Orbit}, \textit{Outward Orbit}, and \textit{Random Walk}, with gradually reduced view overlaps.}
    % \vspace{-3mm}
    \begin{tabular}{lcccc}
        \toprule
        Method & PSNR $\uparrow$ & SSIM $\uparrow$ & LPIPS $\downarrow$ & FID $\downarrow$ \tabularnewline
        \midrule
        \multicolumn{5}{l}{\textit{Forward}} \tabularnewline
        \addlinespace[0.2em]
        \method (w/o layout) & 11.47 & 0.49 & 0.59 & 67.96 \tabularnewline
        \method (w/ layout) & \textbf{17.59} & \textbf{0.69} & \textbf{0.32} & \textbf{34.98} \tabularnewline
        \midrule
        \multicolumn{5}{l}{\textit{Inward orbit}} \tabularnewline
        \addlinespace[0.2em]
        \method (w/o layout) & 12.57 & 0.48 & 0.54 & 64.14 \tabularnewline
        \method (w/ layout) & \textbf{17.30} & \textbf{0.66} & \textbf{0.33} & \textbf{35.57} \tabularnewline
        \midrule
        \multicolumn{5}{l}{\textit{Outward orbit}} \tabularnewline
        \addlinespace[0.2em]
        \method (w/o layout) & 11.14 & 0.60 & 0.47 & 76.73 \tabularnewline
        \method (w/ layout) & \textbf{13.32} & \textbf{0.59} & \textbf{0.46} & \textbf{57.76} \tabularnewline
        \midrule
        \multicolumn{5}{l}{\textit{Random walk}} \tabularnewline
        \addlinespace[0.2em]
        \method (w/o layout) & 11.26 & 0.45 & 0.59 & 98.42 \tabularnewline
        \method (w/ layout) & \textbf{14.07} & \textbf{0.62} & \textbf{0.45} & \textbf{52.10} \tabularnewline
        \bottomrule
    \end{tabular}
    \label{tab:supp:comparison_nvs}
    % \vspace{-3mm}
\end{table}

\subsection{Image to 3D Scene Generation}
\label{sec:supp:img2scene}

\begin{figure*}[t]
    \centering
    \includegraphics[width=\linewidth]{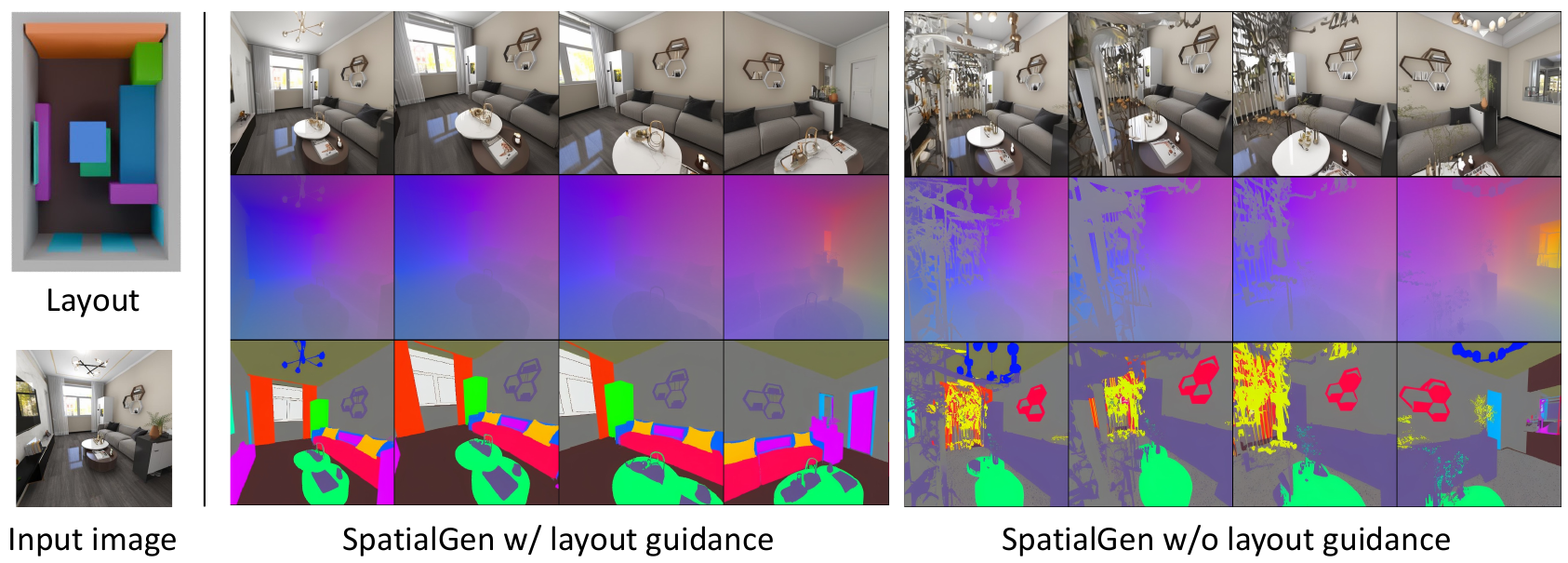}
    \vspace{-6mm}
    \caption{Qualitative comparison of image-to-3D scene generation on our dataset. Given a single input image, our method with layout guidance consistently generates better color images, scene coordinate maps, and semantic maps. }
    \label{fig:image2scene}
    \vspace{-2mm}
\end{figure*}

In this section, we conduct additional image-to-3D scene generation experiments with a focus on two key aspects: (i) generation capability -- the ability to synthesize missing regions for large viewpoint changes; (ii) semantic consistency -- the ability to produce semantically consistent views aligned with the 3D scene layout. Given the lack of accessible literature on the layout-conditioned image-to-3D scene generation task, we compare our multi-view generation model against the version without utilizing layout priors. We employ PSNR, SSIM~\cite{SSIM}, LPIPS~\cite{LPIPS}, and FID~\cite{TTUR} to evaluate the quality of image generation. 

\paragraph{Quantitative Results} \cref{tab:supp:comparison_nvs} reports quantitative results of our method under different camera trajectories. Under all trajectories, the semantic layout improves the results across all metrics. 
Furthermore, the improved FID shows that our method with layout guidance can capture the underlying data distribution more effectively. These results collectively underscore the critical role of incorporating 3D layout information in novel view synthesis.

\paragraph{Qualitative Results} \cref{fig:image2scene} shows example outputs of our method, including RGB images, scene coordinate maps, and semantic maps. Removing the layout input leads to severe artifacts in occluded regions, revealing the limitations of image diffusion models in capturing 3D scene structures. In addition, the semantic map contains unknown content, suggesting degraded semantic prediction without layout input. In contrast, our method with layout guidance generates better novel view images and achieves more reasonable semantic and geometric predictions.

\begin{figure*}[t]
    \centering
    \includegraphics[width=\linewidth]{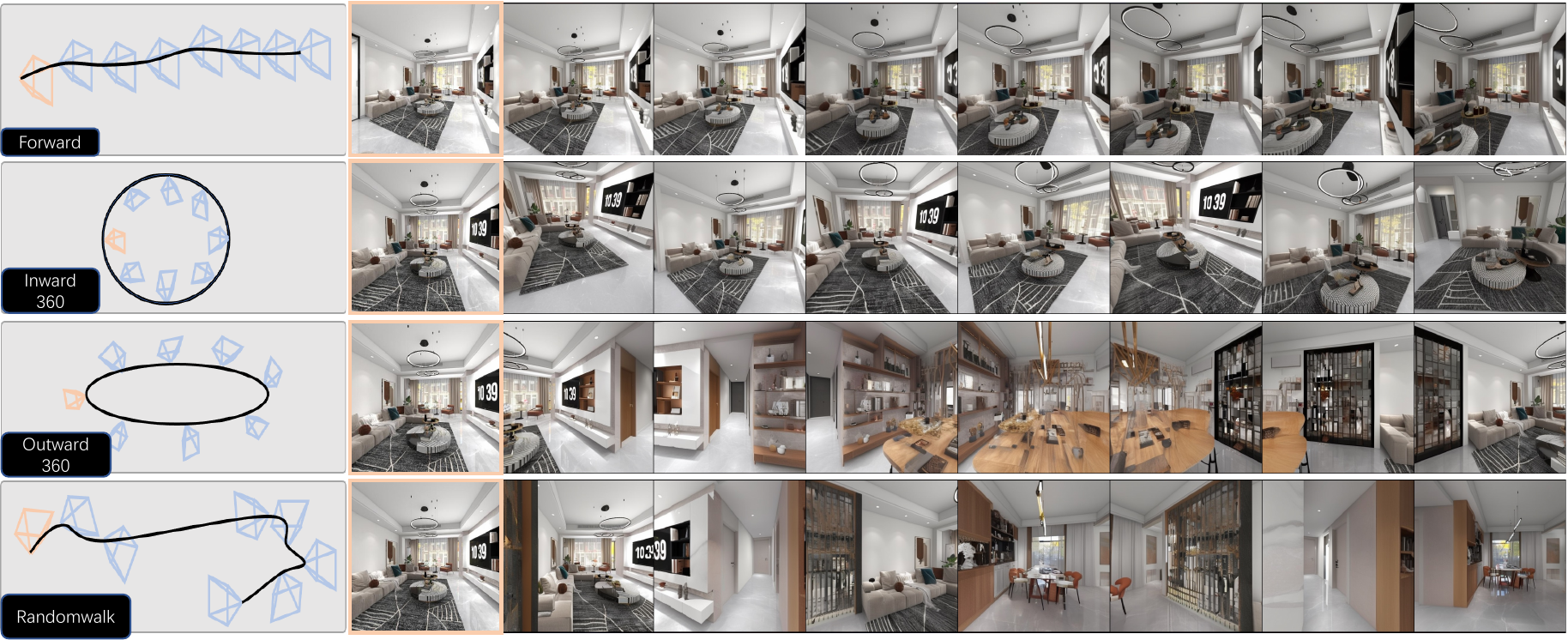}
    \caption{Qualitative results on \method dataset under various camera trajectories. From left to right: input view and target views. \textit{First Row (forward)}: sampled views follow a progressive forward-moving path. \textit{Second row (inward orbit)}: views are directed toward the center of the room, ensuring substantial overlap between them. \textit{Third row (outward orbit)}: views are positioned at the center of the room, looking outward, with an angle of less than $45^\circ$ between two adjacent views. \textit{Bottom (random walk)}: views are selected from a continuous random-walk camera trajectory, producing aggressive viewpoint changes.}
    \label{fig:supp:imgs2scene_different_traj}
\end{figure*}

In \cref{fig:supp:imgs2scene_different_traj}, given a reference image (highlighted in orange box) across four diverse camera trajectories , our method successfully generates 3D-consistent novel views and synthesizes semantically plausible content for areas beyond the original input view.

\subsection{Video-to-3D Scene Generation}
\label{sec:supp:video2scene}

\begin{figure*}[t]
    \centering
    \includegraphics[width=0.8\linewidth]{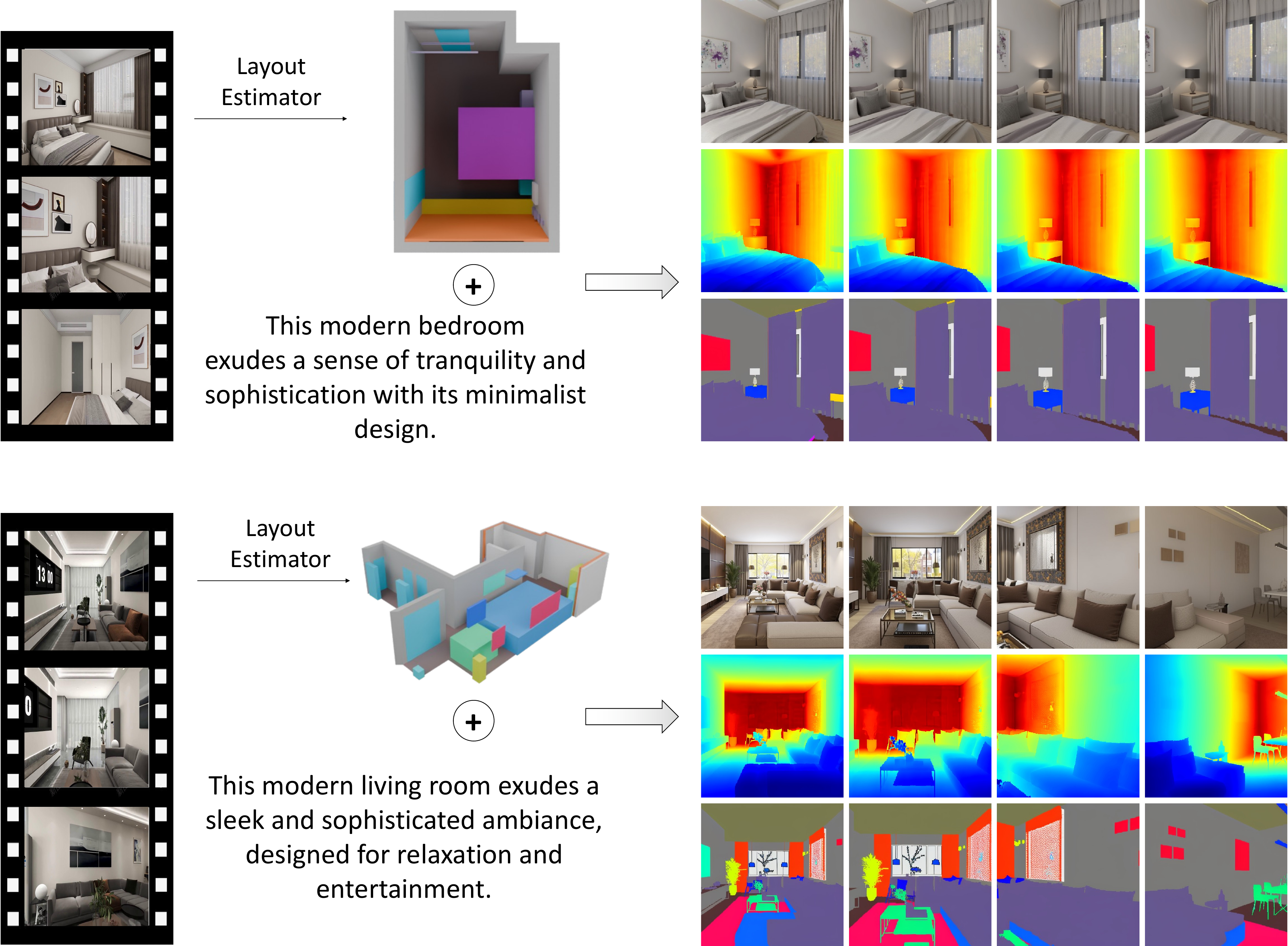}
    \caption{Video-to-New-3D Scene Generation on the SpatialLM Test set~\cite{SpatialLM}. By leveraging the state-of-the-art scene layout estimation method, SpatialLM~\cite{SpatialLM}, we get the reconstructed 3D layout from the video. Then, we perform text-to-3D scene generation conditioned on this layout and additional user-provided text prompts. For clearer visualization of 3D consistency and multi-modal prediction capabilities, we put depth maps here instead of displaying the coordinate maps directly.}
    \label{fig:supp:spatiallm}
\end{figure*}

Given the fact that a well-defined 3D layout is not easy to obtain,
 we apply \method to the challenging task of generating novel 3D scenes from videos. By leveraging a state-of-the-art layout estimation model, SpatialLM~\cite{SpatialLM}, we get the reconstructed 3D layout from the video. Then, we perform text-to-3D scene generation conditioned on this layout and additional user-provided text prompts. This approach allows us to generate entirely new scenes that preserve the structural layout of the original video while altering its stylistic and semantic content based on the text description. We validate this video-to-new-scenes application on the SpatialLM test set. \cref{fig:supp:spatiallm} shows qualitative results.

\subsection{Failure Cases}
\begin{figure}
    \centering
    \includegraphics[width=\linewidth]{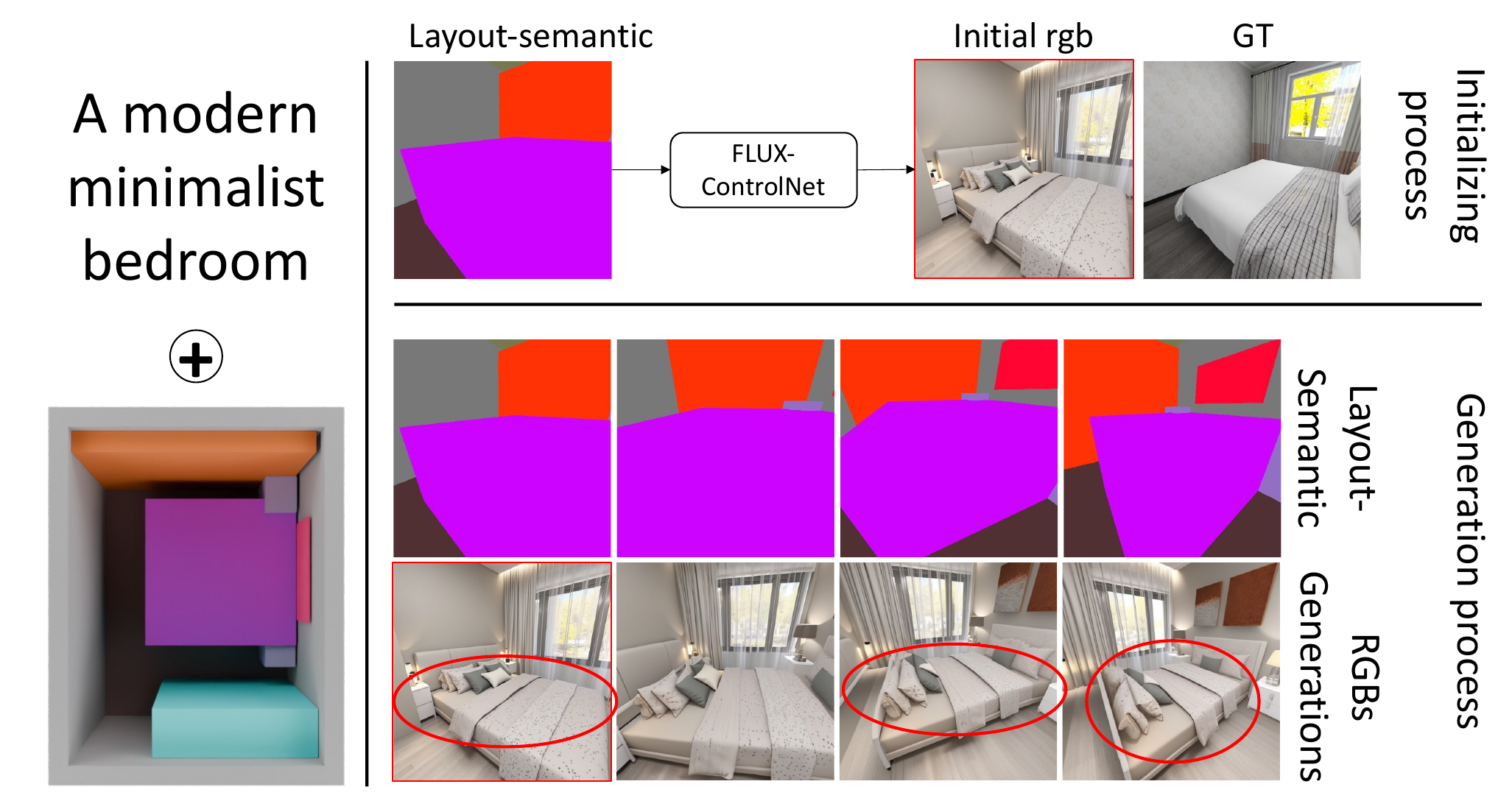}
    \caption{Failure case analysis. The initial RGB image, generated from a layout and text, misorients the bed (red, vs. ground truth). This initial error causes subsequent generations to conflict with the layout.}
    \label{fig:supp:failure_case}
\end{figure}

Because \method relies on an initial RGB image, if the images happen to be inconsistent with the given 3D layout, then it fails to generate consistent results. In this example of text-to-3D scene generation in \cref{fig:supp:failure_case}, we first convert the provided layout and textual description into an initial RGB image. However, the initialization process yields an RGB output (highlighted in red) that does not align with the given layout—specifically, the orientation of the bed is entirely reversed compared to the ground truth. Consequently, during the subsequent generation stage, the resulting outputs exhibit inconsistencies and fail to align properly with both the layout and the initial RGB image.

\section{Ablation Studies}
\label{sec:supp:ablation}

In this section, we conduct additional ablation studies to verify the layout guidance, design choice of network architecture, and the number of input views. 

\begin{figure}
    \centering
    \includegraphics[width=\linewidth]{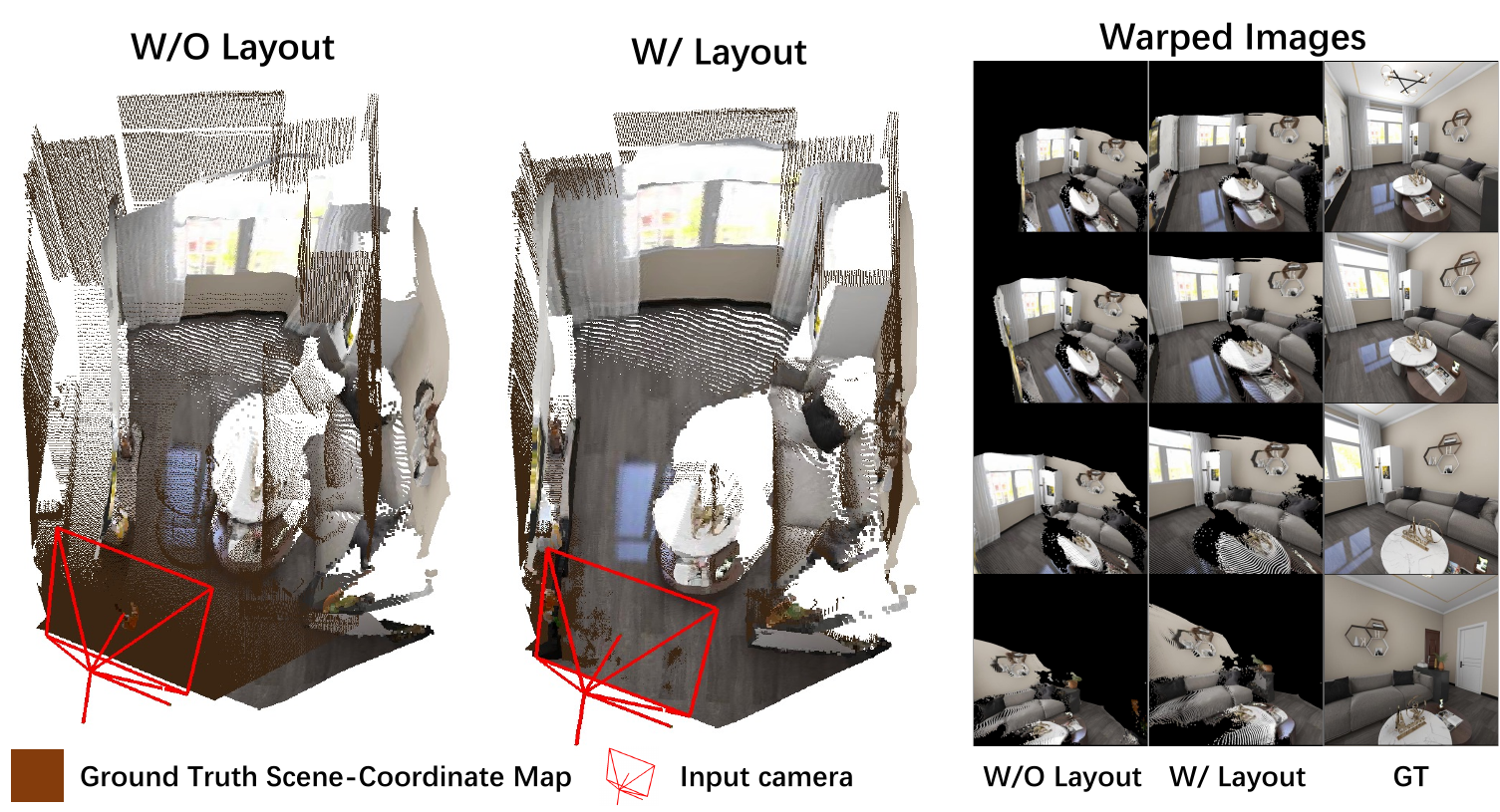}
    \caption{Comparing geometric prediction quality between our \textit{(W/ layout)} and \textit{(W/O layout)}. The first two columns show the predicted scene coordinate maps,  where our method (W/ layout) achieves better alignment with ground-truth geometry (brown color point cloud) compared to the counterpart without layout guidance (W/O layout). Correspondingly, the warped images projected by the predicted scene coordinates demonstrate improved spatial consistency and reduced artifacts.}
    \label{fig:analyze_image2scene}
\end{figure} 

\begin{figure*}
    \centering
    \includegraphics[width=\linewidth]{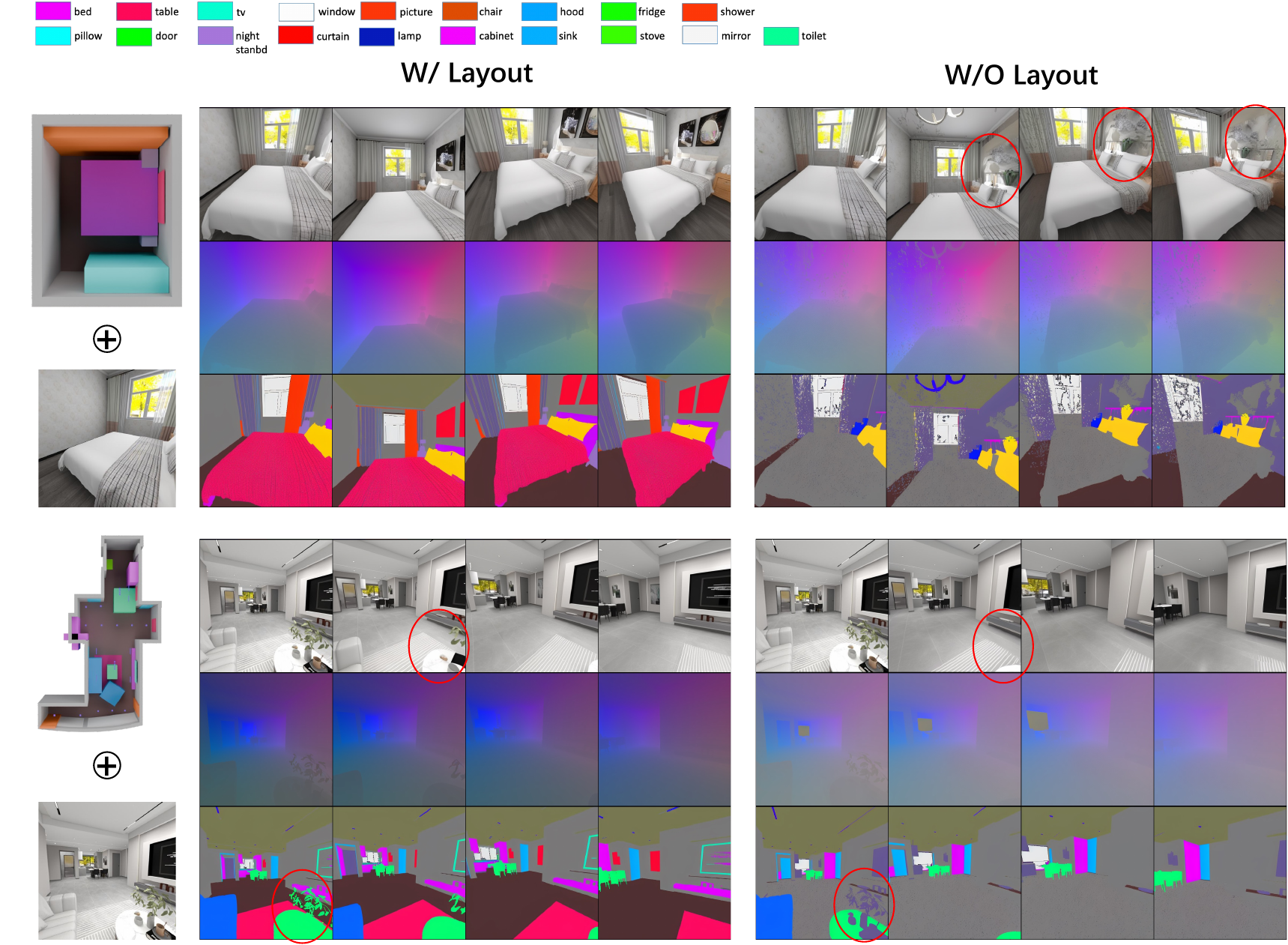}
    \caption{Ablation on the effectiveness of layout as guidance.}
    \label{fig:supp:abla_layout_img2scene}
\end{figure*}

\begin{figure*}
    \centering
    \includegraphics[width=\linewidth]{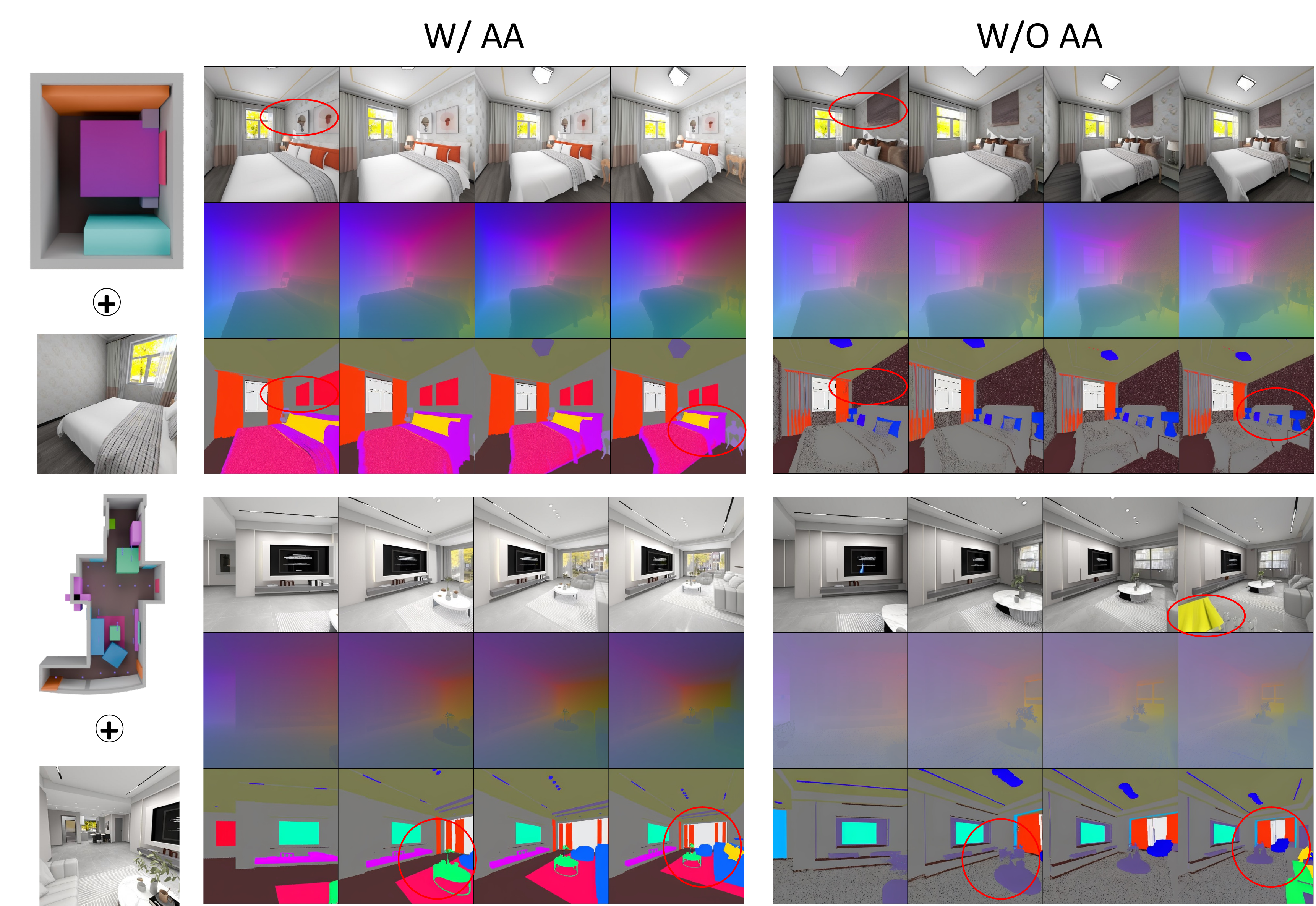}
    \caption{Ablation on the effectiveness of the Multi-view Multi-modal Alternating Attention mechanism.}
    \label{fig:supp:abla_aa_img2scene}
\end{figure*}

\paragraph{Ablation on layout guidance} We first study the effect of layout guidance to validate our design. We compare our full model (denoted as \textit{W/ Layout}) against a variant that removes layout priors (denoted as \textit{W/O Layout}). The \textit{W/O Layout} variant is implemented similarly to CAT3D~\cite{CAT3D} but incorporates our multi-view multi-modal alternating attention module to enable multi-modal output. Both models are trained identically for single-image 3D scene generation.

\Cref{fig:supp:abla_layout_img2scene} presents a faithful comparison, showing generated RGB outputs, scene coordinate maps, and semantic maps (top to bottom) from a given input (left-most column). As the red circles highlight, the \textit{W/O Layout} variant produces artifacts in occluded regions, exhibits imperfect image-pose alignment, and generates degraded dense predictions. These failures indicate the inherent limitations of relying solely on image diffusion priors for 3D scene generation. In contrast, our full model \textit{W/ Layout} leverages explicit layout guidance to achieve superior novel-view synthesis and more accurate geometry and semantic predictions.

Furthermore, \cref{fig:analyze_image2scene} provides an in-depth analysis of 3D consistency. We visualize the predicted scene coordinates for the input view, demonstrating that the \textit{W/ layout} predictions achieve better alignment with the ground truth. This superior alignment provides more accurate warped images for all target viewpoints, explaining its clear superiority over the \textit{W/O layout} baseline.

\paragraph{Ablation on Alternating Attention mechanism} 
We further evaluate the design choice of Multi-view Multi-modal Alternating Attention (AA). We compare our full model (denoted as \textit{W/ AA}) against a variant that disables the multi-modal attention (denoted as \textit{W/O AA}). Both models share the same training protocol as the aforementioned.

Qualitative results in \Cref{fig:supp:abla_aa_img2scene} reveal a critical weakness in the ablated model. As indicated by the red circles, the \textit{W/O AA} model fails to produce semantically meaningful segmentations. This deficiency corrupts the geometry in the scene coordinate maps and resulting in less coherent multi-view RGB generation. This demonstrates that without an explicit mechanism to align and refine information across modalities (RGB, geometry, semantics), the model cannot effectively leverage their synergies.

Our complete model \textbf{W/ AA} directly addresses this limitation. By facilitating cross-modal interaction, the AA mechanism enables more precise semantic labels and geometrically consistent scene coordinates. This improvement subsequently elevates the fidelity and view-consistency of the final RGB outputs, confirming that the alternating attention is pivotal for high-quality, multi-modal scene generation.

\paragraph{Ablation on number of input views} In \cref{tab:ab:number}, we evaluated \method using different numbers of input views in the \textit{inward orbit} camera configuration. Increasing the number of input views enhances all metrics, particularly the FID score; this implies that a greater input views enhances semantic consistency.

\begin{table}
    \centering
    \small
    \renewcommand{\tabcolsep}{4pt}
    \caption{Effect of the number of input views in the inward orbit setting.}
    \begin{tabular}{ccccc}
        \toprule
        \#input views & PSNR $\uparrow$ & SSIM $\uparrow$ & LPIPS $\downarrow$ & FID $\downarrow$ \tabularnewline
        \midrule
        1 & 17.30 & 0.66 & 0.33 & 35.57 \tabularnewline
        3 & 17.83 & 0.67 & 0.31 & 28.72 \tabularnewline
        6 & \textbf{18.33} & \textbf{0.67} & \textbf{0.31} & \textbf{21.93} \tabularnewline
        \bottomrule
    \end{tabular}
    \label{tab:ab:number}
\end{table}

% {
% \small
% \bibliographystyle{ieeenat_fullname}
% \bibliography{spatialgen}
% }
\end{document}